\documentclass{article} 
\usepackage{iclr2026_conference,times}
\usepackage{booktabs}
\usepackage{multirow}
\usepackage{geometry}
\usepackage{amsmath}
\usepackage{caption}
\usepackage{xcolor}

\usepackage{amsmath,amsfonts,bm}

\def\eqref#1{equation~\ref{#1}}

\def\1{\bm{1}}

\DeclareMathAlphabet{\mathsfit}{\encodingdefault}{\sfdefault}{m}{sl}
\SetMathAlphabet{\mathsfit}{bold}{\encodingdefault}{\sfdefault}{bx}{n}

\usepackage{hyperref}
\usepackage{url}
\usepackage{siunitx}   
\usepackage{graphicx}   
\usepackage{adjustbox}
\usepackage{pifont}
\usepackage{fontawesome}
\usepackage{wasysym}
\usepackage{marvosym}
\usepackage{listings}
\usepackage[table]{xcolor}
\usepackage{tcolorbox}
\usepackage{enumitem}

\iclrfinalcopy
\title{EgoNight: Towards Egocentric Vision Understanding at Night with a Challenging Benchmark}

\author{ \small Deheng Zhang$^{1}\thanks{\ means equal contribution; † denotes corresponding author.}$\ , \small Yuqian Fu$^{1 * \dagger}$, \small Runyi Yang$^{1}$, \small Yang Miao$^{1}$, \small Tianwen Qian$^{2}$, \small Xu Zheng$^{1,3}$, \small \textbf{Guolei Sun}$^{4}$, \\ \small \textbf{Ajad Chhatkuli}$^{1}$,  \small \textbf{Xuanjing Huang}$^{5}$, \small \textbf{Yu-Gang Jiang}$^{5}$, \small \textbf{Luc Van Gool}$^{1}$, \small \textbf{Danda Pani Paudel}$^{1}$\\ $^{1}$\small INSAIT, Sofia University “St. Kliment Ohridski” \quad $^{2}$\small East China Normal University \\ $^{3}$\small HKUST(GZ) \quad $^{4}$\small Nankai University \quad $^{5}$\small Fudan University }

\begin{document}

\maketitle

\begin{figure}[h]
    \centering
    \vspace{-0.3in}
    \includegraphics[width=0.9\linewidth]{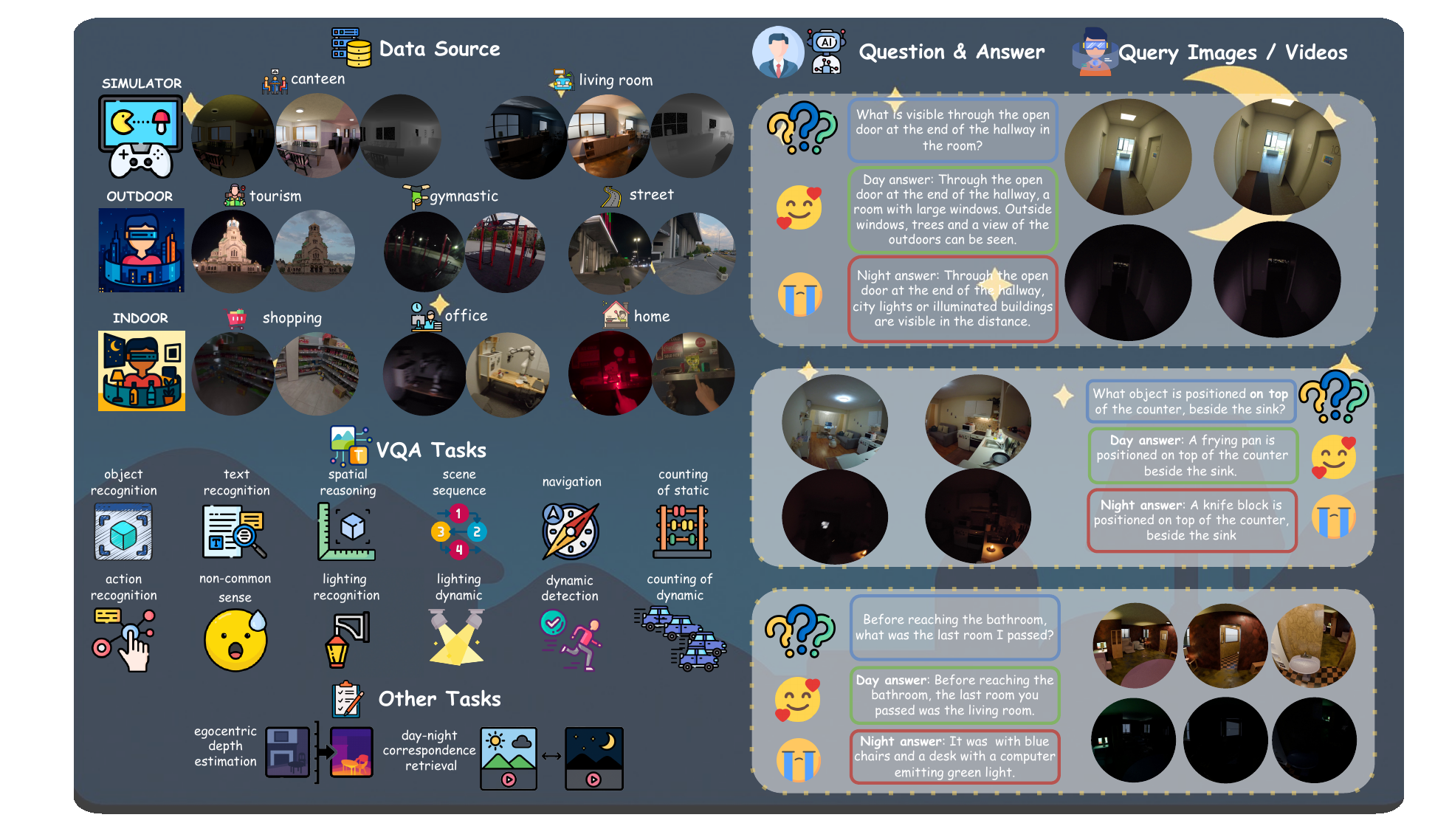}
    \caption{\textbf{Overview of the EgoNight.}  
    EgoNight integrates diverse video sources spanning synthetic environments, real-world indoor and outdoor scenes, recorded under both daytime and nighttime conditions, with spatial and temporal alignment.  
    It consists of three benchmarks: (i) \textit{egocentric VQA} as the primary focus, (ii) \textit{day–night correspondence retrieval}, and (iii) \textit{egocentric depth estimation}, all targeting the challenges of low-light egocentric vision.  The day–night alignment (illustrated on the right with VQA examples) enables rigorous analysis of illumination gaps in MLLMs.
    }
    \label{fig:teaser}
\end{figure}

\begin{abstract}
Most existing benchmarks for understanding egocentric vision focus primarily on daytime scenarios, overlooking the low-light conditions that are inevitable in real-world applications. To investigate this gap, we present \emph{EgoNight}, the first comprehensive benchmark for nighttime egocentric vision, with visual question answering (VQA) as the core task. A key feature of EgoNight is the introduction of day–night aligned videos, which enhance night annotation quality using the daytime data and reveal clear performance gaps between lighting conditions. To achieve this, we collect both synthetic videos rendered by Blender and real-world recordings, ensuring that scenes and actions are visually and temporally aligned. Leveraging these paired videos, we construct \emph{EgoNight-VQA}, supported by a novel day-augmented night auto-labeling engine and refinement through extensive human verification. Each QA pair is double-checked by annotators for reliability. In total, EgoNight-VQA contains 3658 QA pairs across 90 videos, spanning 12 diverse QA types, with more than 300 hours of human work. Evaluations of the state-of-the-art multimodal large language models (MLLMs) reveal substantial performance drops when transferring from day to night, underscoring the challenges of
reasoning under low-light conditions. Beyond VQA, EgoNight also introduces two auxiliary tasks, day–night correspondence retrieval and egocentric depth estimation at night, that further explore the boundaries of existing models. We believe EgoNight-VQA provides a strong foundation for advancing application-driven egocentric vision research and for developing models that generalize across illumination domains The code and data can be found in \href{https://github.com/dehezhang2/EgoNight}{https://github.com/dehezhang2/EgoNight}. 
\end{abstract}

\section{Introduction}
With the rapid development of wearable devices, egocentric vision understanding has become increasingly important. Unlike third-person vision, egocentric perception naturally aligns with the way humans perceive, understand, and interact with the world. A robust egocentric vision system can not only serve as an intelligent assistant in daily activities~\cite{yang2025egolife} but also play a crucial role in embodied AI and robotic learning~\cite{li2025clivis, kareer2025egomimic}. Beyond these general applications, egocentric vision holds unique potential for assisting specific user groups such as people who are blind or visually impaired~\cite{xiao2025egoblind}, or physically disabled~\cite{zhang2023accessiblerobotcontrolmixed}, enabling technologies that enhance navigation, accessibility, and real-time scene understanding.

Significant efforts have been made to advance egocentric vision understanding, including the construction of large-scale ego-centric datasets such as EPIC-KITCHENS~\cite{damen2020epic}, Ego4D~\cite{grauman2022ego4d}, and Ego-Exo4D~\cite{grauman2024ego}; the design of diverse and challenging benchmarks such as EgoTaskQA~\cite{jia2022egotaskqa}, EgoSchema~\cite{mangalam2023egoschema}, and EgoTempo~\cite{plizzari2025omnia}; and the development of egocentric multimodal large language models (MLLMs) such as EgoVLPv2~\cite{pramanick2023egovlpv2}, EgoGPT~\cite{yang2025egolife}, and Exo2Ego~\cite{zhang2025exo2ego}. Despite these advances, almost all prior works focus on daytime scenarios with favorable lighting. 
In contrast, real-world egocentric systems, for example, intelligent personal assistants for navigation, must operate at night, under low light, uneven illumination, and severely limited visibility. This motivates us to investigate egocentric vision at night, focusing on complex scene understanding and reasoning tasks.

A central challenge in constructing such a benchmark lies in obtaining suitable video sources that capture the characteristics of nighttime environments and developing annotation methods that ensure high labeling quality. To address this, we place particular emphasis on \textit{day–night aligned videos}, which not only allow us to leverage daytime data to annotate nighttime videos, but also enable rigorous performance comparisons across day and night lighting conditions.
However, in practice, collecting perfectly aligned day–night pairs in the real world is highly non-trivial. To overcome this, we leverage Blender~\cite{iraci2013blender}, where scene layouts, camera trajectories, and lighting can be precisely controlled, enabling the synthesis of the desired videos. This produces EgoNight-Synthetic, a collection of 50 ideally aligned egocentric pairs spanning diverse and complex indoor scenarios with varying illumination levels. To complement synthetic data with real-world evidence, we design a \textit{video-guided recording protocol} to construct \textbf{EgoNight-Sofia}, which contains 20 pairs of real-world egocentric videos with spatially and temporally aligned day–night counterparts. These videos cover realistic use cases (e.g., “Where did I put my keys?”, “How much is the item I saw in the grocery shop?”), spanning both indoor and outdoor environments under diverse illumination sources such as streetlights, flashlights, and candles. 
Finally, we incorporate 20 nighttime videos from the Oxford Day-and-Night dataset~\cite{wang2025seeing}, termed \textbf{EgoNight-Oxford}, which serve as an additional testbed despite lacking day-night alignment. Together, these three video sources constitute our \textbf{EgoNight} dataset, which is the first egocentric dataset providing day–night aligned correspondences, as mainly summarized in Fig.~\ref{fig:teaser}.

The videos in EgoNight pave the way for constructing challenging benchmarks to evaluate the capabilities of existing models. Among many egocentric tasks, we focus on the egocentric video question answering, a flagship task that best reflects high-level understanding in egocentric vision. 
Specifically, to comprehensively evaluate model abilities, we first propose a diverse set of QA types, spanning well-studied tasks (e.g., object recognition, spatial reasoning, action recognition, counting, text recognition) as well as several underexplored dimensions (e.g., temporal scene sequence understanding, navigation, lighting recognition, and non–common-sense reasoning). These are further organized into paired and unpaired QA types, depending on whether day–night counterparts share the same questions and answers. To construct the benchmark at scale, we then develop a \textit{novel three-stage day-augmented auto-labeling pipeline} that leverages daytime videos to assist in generating question–answer pairs for nighttime clips, followed by extensive human verification to ensure accuracy and reliability. Building EgoNight and annotating VQA required \textbf{over 300 hours of human effort}, with each QA pair verified by at least one expert annotator. This process results in the high-quality \textbf{EgoNight-VQA} dataset, comprising 3,658 QA pairs. Beyond VQA, we introduce two auxiliary tasks with dedicated testbeds: day–night correspondence retrieval, which evaluates cross-illumination matching, and egocentric depth estimation at night, which is crucial for navigation and interaction in embodied AI. These two tasks further broaden the benchmark and expose new challenges for existing models.
 
Our extensive experiments across three video sources, three tasks, and 10 state-of-the-art multimodal large language models reveal that nearly all models (including closed-source models such as GPT and Gemini) struggle on this challenging benchmark, with a clear and consistent performance gap between day and night. This highlights the unsolved challenges of egocentric vision at nighttime and calls for more robust models that generalize across illumination conditions. 
Besides, we highlight that our newly introduced QA types, covering lighting recognition/dynamic, scene sequence reasoning, navigation, and non–common-sense reasoning, are substantially more challenging than well-studied categories, revealing fresh difficulties for MLLMs. We further prove synthetic data is highly correlated with real data and effectively boosts real-world performance through fine-tuning. Our pilot studies further show that fine-tuning on specialized subset of data improves model performance through adapting vision encoder into low-light domain and aligning the language model to the uncertain features during night.

Our main contributions are threefold: i) \textbf{EgoNight Dataset}: We present the first egocentric dataset that systematically addresses nighttime conditions, featuring day–night aligned videos from synthetic (EgoNight-Synthetic), real-world (EgoNight-Sofia), and existing (EgoNight-Oxford) sources. ii) \textbf{Benchmark Suite}: We build a comprehensive benchmark centered on egocentric VQA with diverse QA types and 3658 fully human-verified QA pairs, complemented by egocentric depth estimation at night and day--night correspondence retrieval tasks. iii) \textbf{Empirical Insights}: Extensive evaluations reveal clear day–night performance gaps, underscoring illumination robustness as a key challenge; our newly proposed QA types are also validated to pose practical difficulties for current MLLMs.

\section{Related Works}
\subsection{Egocentric Datasets and VQA Benchmarks}
A series of large-scale egocentric datasets, such as EPIC-KITCHENS~\cite{damen2020epic}, Ego4D~\cite{grauman2022ego4d}, Ego-Exo4D~\cite{grauman2024ego}, and EgoExoLearn~\cite{huang2024egoexolearn}, have laid the foundation for a wide range of tasks, including action recognition~\cite{sudhakaran2019lsta}, object detection~\cite{ren2010figure}, pose estimation~\cite{luo2021dynamics}, video generation~\cite{liu2021cross}, Ego-Exo correspondence~\cite{fu2024objectrelator,pan2025v} and Ego-Exo translation~\cite{liu2024exocentric, mahdi2025exo2egosyn}. Among these, we are particularly interested in egocentric visual question answering (VQA)~\cite{fan2019egovqa}, which provides a natural and human-like framework for comprehensively evaluating model performance through question–answer interactions. In recent years, several egocentric VQA benchmarks have been proposed, including EgoVQA~\cite{fan2019egovqa}, EgoTaskQA~\cite{jia2022egotaskqa}, EgoSchema~\cite{mangalam2023egoschema},  EgoThink~\cite{cheng2024egothink}, EgoTempo~\cite{plizzari2025omnia}, EgoCross~\cite{li2025egocross}, EgoBlind~\cite{xiao2025egoblind}, EgoMemoria~\cite{ye2024mm}, HourVideo~\cite{chandrasegaran2024hourvideo}, EgoLifeQA~\cite{yang2025egolife} with different focuses. 
However, nearly all of them are confined to daytime or well-lit scenarios, leaving model performance in low-light or nighttime conditions largely unexplored. 
The Oxford Day-and-Night dataset~\cite{wang2025seeing} is a partial exception but was not designed for VQA and lacks day–night alignment. 
This makes EgoNight and EgoNight-VQA fundamentally distinct from prior benchmarks.

\subsection{MLLMs for Video Understanding}
The rapid development of multimodal large language models (MLLMs) has substantially advanced the frontier of video understanding. Prominent open-source models include Qwen-VL~\cite{Qwen-VL}, InternVL~\cite{chen2024internvl}, Video-LLaMA~\cite{zhang2023video}, LLaVA-NeXT-Video~\cite{li2024llava}, and GLM-V~\cite{hong2025glm}, while closed-source commercial systems such as GPT-4V~\cite{achiam2023gpt} and Gemini~\cite{comanici2025gemini} demonstrate even stronger capabilities in video captioning, summarization, and open-ended visual question answering.
Building on these advances, egocentric MLLMs have emerged to adapt foundation models from exocentric to first-person perspectives. 
Representative examples include EgoVLPv2~\cite{pramanick2023egovlpv2} for improved video–language cross-modal fusion, EgoGPT~\cite{yang2025egolife} fine-tuned with egocentric captioning and QA, MM-Ego~\cite{ye2024mm} with a memory mechanism for long videos, and Exo2Ego~\cite{zhang2025exo2ego} leveraging exocentric data for egocentric generalization. These works highlight the potential of MLLMs as egocentric assistants. However, nearly all of them are developed and tested under well-lit daytime conditions, leaving their robustness in low-light or nighttime scenarios unexplored.
 Nevertheless, almost all existing MLLMs are developed and evaluated under well-lit daytime conditions, with little consideration of low-light or nighttime videos, leaving their robustness in low-light or nighttime scenarios unexplored.

\subsection{Cross-Domain Generalization}
Domain generalization~\cite{zhou2022domain} is a long-standing challenge in computer vision, where models trained on one distribution must adapt to another. Shifts can arise from semantic drift, style changes, or variations in weather and lighting. Many algorithms have been validated across tasks such as image classification~\cite{li2017deeper, zhou2021domain, huang2023evidential, peng2024advancing, peng2026mitigating}, object detection~\cite{fu2024cross, li2025domain}, action recognition~\cite{pan2020adversarial, fu2020depth, bian2011cross, zou2021annotation, zou2020adaptation}, few-shot learning~\cite{guo2020broader, fu2021meta, fu2023styleadv, zhang2025decoupling}, and autonomous driving~\cite{li2022cross, li2023domain}.
In contrast, cross-domain transfer for MLLMs, especially in video understanding, remains underexplored, with only a few recent attempts (e.g., CL-CrossVQA~\cite{zhang2025cl}, VQA-GEN~\cite{unni2023vqa}, Super-CLEVR~\cite{li2023super}). 
However, none of them are targeted for egocentric video, which is naturally different from exocentric videos in terms of recorded images, camera motion, and contained information. The most relevant effort to us is EgoCross~\cite{li2025egocross}, an egocentric VQA benchmark that moves beyond daily activities to evaluate model generalization across distinct long-tail, specialized domains such as surgery and industrial settings. In this paper, however, we investigate MLLMs from a different perspective, robustness under nighttime conditions, a common and ubiquitous scenario in daily life, yet previously overlooked dimension of domain generalization in egocentric video understanding.

\section{EgoNight Dataset \& Benchmarks}
\subsection{Video Source Collection}~\label{sec:video_source}
\vspace{-2em}
\par
\begin{figure}[htbp]
    \vspace{-0.2in}
    \centering
    \includegraphics[width=\linewidth]{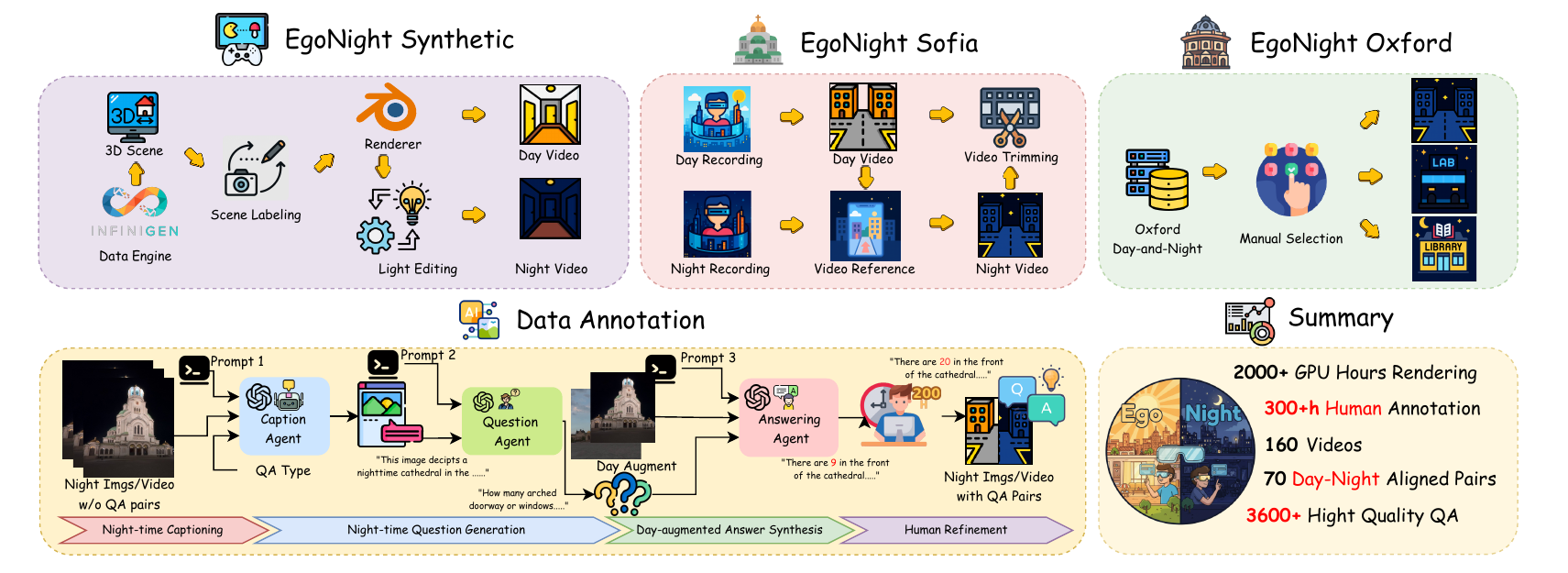}
    \vspace{-0.1in}
    \caption{\textbf{EgoNight construction and EgoNight-VQA annotation.} EgoNight integrates EgoNight-Synthetic, EgoNight-Sofia, and EgoNight-Oxford sources. Annotation is achieved via a novel three-stage day-augmented Auto QA generation pipeline with 300+ hours of human refinement, resulting in over 3600 high-quality QA pairs.
    }
    \label{fig:main_pipeline}
    \vspace{-0.2in}
\end{figure}

\noindent\textbf{Overview \& Design Principles.} 
EgoNight is built to systematically evaluate MLLMs under challenging nighttime conditions, which are critical for developing robust intelligent assistants. 
The collection of video sources follows four principles: 
\ding{172} \textit{Reflect real-world challenges}, such as walking on dimly lit streets or navigating indoors during power outages; 
\ding{173}  Involve \textit{natural camera movements} and preferably capture \textit{actions and interactions} with the environment to evaluate both static perception and dynamic understanding; 
\ding{174} Ensure \textit{diversity of scenarios, illumination, and task difficulties}, spanning indoor, outdoor, office, and grocery settings, lighting from streetlights, flashlights, headlights, and candles, and task levels from easy (relatively clear), through medium (partially visible), to hard (barely visible).
\ding{175} Enable rigorous analysis through \textit{day–night paired videos}, where scenes, trajectories, and actions remain consistent across conditions so that differences can be attributed solely to illumination.
To meet these requirements, EgoNight integrates three complementary video sources, as illustrated in the upper part of Fig.~\ref{fig:main_pipeline} and detailed below.

\noindent\textbf{EgoNight-Synthetic.} 
To obtain perfectly aligned day–night video pairs, we used a simulation environment where every element can be precisely controlled, including the scene layout, camera path, and lighting. This ensures that the day and night videos match exactly at the pixel and frame level, with lighting being the only difference. We first employ Infinigen \cite{infinigen2023infinite} to generate diverse indoor 3D scenes. Human annotators cleaned and refined these scenes, then simulated walking through the space at a normal speed (1.2 m/s), recording the camera trajectory. We replayed the same trajectory under different lighting conditions. Daytime videos were rendered using Blender \cite{iraci2013blender}, and we adjusted the lighting to create the corresponding nighttime versions.

In total, EgoNight-Synthetic contains 50 pairs of egocentric videos, covering more than 100 environment assets. These include indoor scenes such as kitchens, bathrooms, and living rooms, populated with over 50 diverse object categories (e.g., windows, tables, beds, chairs, lamps, bookshelves, plates). 
We design multiple illumination setups, ranging from uniformly lit rooms to sparsely localized lighting, and incorporate three difficulty levels with a different range of motion blur, sensor noise, and illumination level. Besides RGB frames, Blender also allows us to generate ground-truth depth and normals (see Appendix Sec.~\ref{sec:more_construction}), making EgoNight-Synthetic richer and more versatile.

\noindent\textbf{EgoNight-Sofia.} 
To include realistic human–environment interactions missing from synthetic videos, we also recorded our own day–night paired videos. Capturing perfectly aligned real-world pairs is challenging, so we designed a practical video-guided recording strategy with post-trimming for better alignment.

We first record a daytime video with an ego-wearer exploring an environment while viewing the live camera feed on a phone screen. For the nighttime version, the same person, device, and viewpoint are kept unchanged. Instead of using live preview, the recorded daytime video is played back on the phone, serving as visual guidance to help the ego-wearer match walking speed, viewpoints, and actions. After brief practice, this approach proved more stable and reliable than methods like using landmarks or memorized trajectories. Post-trimming is applied to further refine spatial and temporal consistency.

Our real-world dataset, EgoNight-Sofia, contains 20 day–night paired videos recorded in Sofia, Bulgaria. Despite its modest size, it is a rare resource capturing diverse real-world everyday scenarios, including apartments, offices, grocery stores, streets, tourist spots, and outdoor fitness areas. The recordings include natural actions such as drinking water, locking doors, placing keys, charging devices, or checking price labels—leading to realistic VQA cases (e.g., “Where did I put my keys?”, “How much was the drink?”, “Did I turn left?”). Illumination sources include street lights, lamps, flashlights, and candles.

\noindent\textbf{EgoNight-Oxford.}  
Oxford Day–Night \cite{wang2025seeing} is a notable exception that also includes egocentric videos captured under both daytime and nighttime conditions. Although it was originally designed for 3D vision tasks such as novel view synthesis, it offers illumination variations across five representative locations in Oxford. However, the day and night videos are not spatially or temporally aligned.

To enrich EgoNight with more realistic nighttime content, particularly for urban outdoor scenes, we manually select 20 nighttime segments to form EgoNight-Oxford, based on two criteria: (i) minimal overlap in trajectories and locations, and (ii) genuinely low-light conditions. These segments serve as a complementary testbed for evaluating model generalization under illumination changes when paired alignment is unavailable.

For both EgoNight-Sofia and EgoNight-Oxford, human annotators categorize each video into easy, medium, or hard levels. Together, these sources provide EgoNight with a balanced combination of precise alignment, natural dynamics, and broad real-world diversity.
\vspace{-1em}

\subsection{EgoNight-VQA Benchmark Reconstruction}~\label{sec:vqa}
\textbf{QA Task Taxonomy.} To thoroughly assess models from multiple perspectives, we define a diverse taxonomy of 12 QA tasks. Some of these categories are well-studied and have been explored in previous egocentric VQA benchmarks, such as object/action/text recognition, counting, and spatial reasoning. 
Others are much less studied or newly proposed in EgoNight-VQA, including scene sequence and navigation (which require not only visual perception but also memory and spatial reasoning), illumination recognition and illumination change (designed to test models’ understanding of lighting concepts), and non–common-sense reasoning (e.g., detecting abnormal cases such as a door inserted into a wall in the synthetic data). More detailed explanations of QA types can be found in the Appendix Sec.~\ref{sec:qa_type_def}. We further organize these categories into \textit{paired} and \textit{unpaired} QA types, depending on whether the same questions can be consistently applied across day–night counterparts:  1)  \textbf{Paired QA Types.} These cover contexts that remain unchanged across day and night, allowing the same QA pairs to be used for both videos and thus providing a clean testbed for measuring performance gaps. 
  Specifically, we include: 
  \ding{172} \textit{object recognition}, 
  \ding{173} \textit{text recognition}, 
  \ding{174} \textit{spatial reasoning}, 
  \ding{175} \textit{scene sequence}, 
  \ding{176} \textit{navigation}, 
  \ding{177} \textit{counting of static}, 
  \ding{178} \textit{action recognition}, and 
  \ding{179} \textit{non–common-sense reasoning}. 
2) \textbf{Unpaired QA Types.} These include categories that are impractical to pair across day and night, or are only meaningful in the nighttime condition. 
  We consider: 
  \ding{172} \textit{lighting recognition}, 
  \ding{173} \textit{lighting dynamic}, 
  \ding{174} \textit{dynamic detection}, and 
  \ding{175} \textit{counting of dynamic}.  We control QA clip duration by task type.
For static or spatial tasks (e.g., object recognition, lighting recognition), we use short clips of 3 seconds to minimize redundancy; For dynamic or temporal tasks (e.g., action recognition, navigation), the entire video is used to capture the complete context. Following recent works~\cite{plizzari2025omnia, xiao2025egoblind}, we adopt the \textit{open-ended QA} setting over the closed-form multiple-choice format, as it better reflects natural human–AI interactions.

\begin{figure}[h]
    \centering
    \vspace{-0.1in}
    \includegraphics[width=\linewidth]{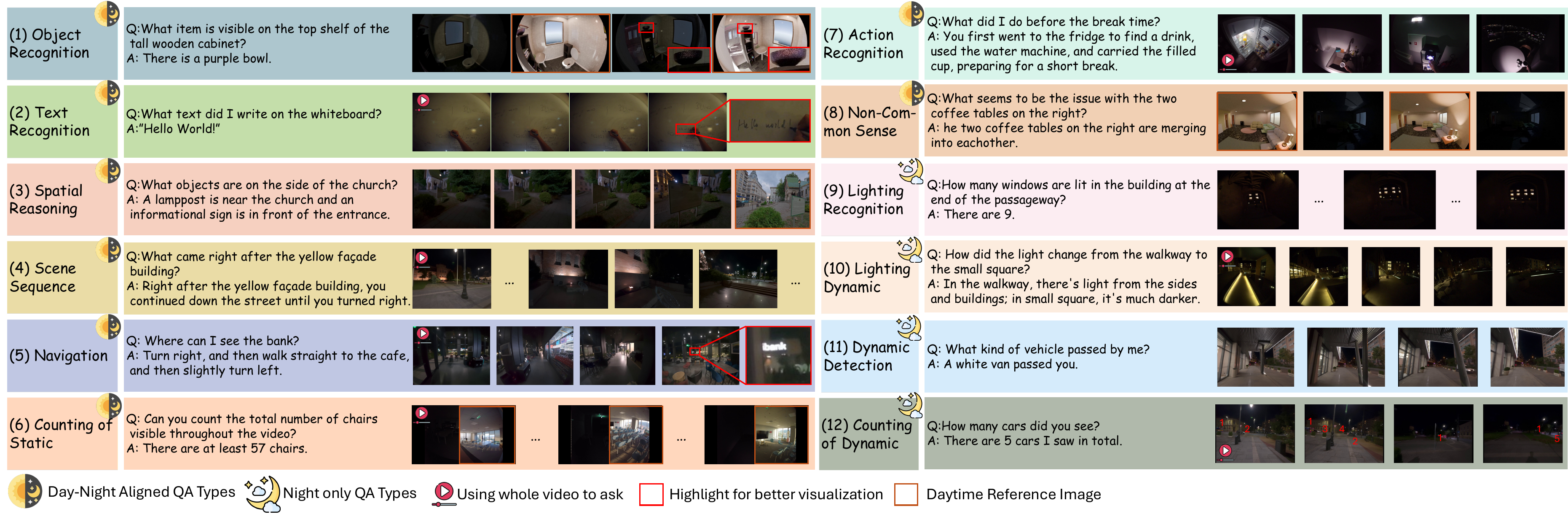}
    \vspace{-0.1in}
    \caption{
     \textbf{QA types with examples.} The first eight are \textit{paired} types, where the same question–answer applies to both day and night clips; the last four are \textit{unpaired}, evaluated only at night. QA Types have various durations, with static or spatial tasks (e.g., 1 and 3) using short clips, while dynamic or temporal tasks (e.g., 4 and 5) use full videos. 
    }
    \label{fig:tasktaxonomy}
      \vspace{-0.1in}
\end{figure}

A detailed summary of each QA type, including whether it is paired or unpaired, clip duration, and example questions, is provided in Fig.~\ref{fig:tasktaxonomy}. This taxonomy makes EgoNight-VQA not only diverse and well-structured but also novel, introducing illumination reasoning and other challenges uniquely tied to nighttime egocentric vision.

\noindent\textbf{Day-Augmented Auto QA Generation.}
Constructing large-scale QA pairs for nighttime videos is particularly challenging due to low visibility, which makes direct annotation both time-consuming and error-prone. 
To address this, as illustrated in the lower part of Fig.~\ref{fig:main_pipeline}, we design a novel three-stage \textit{day-augmented auto QA generation} pipeline that leverages aligned daytime videos as a strong prior for annotating their nighttime counterparts.  

Specifically, the pipeline is tailored to each QA type and consists of three stages:  

1) \textbf{Nighttime captioning.}   
For each clip, we prompt advanced MLLMs to generate detailed captions with an explicit focus on the target QA type (e.g., highlighting object-related attributes for object recognition or text/logos for text recognition). This ensures that the captions capture the most key information or construct relevant QA pairs.

2) \textbf{Nighttime question generation.}  
The caption, together with the corresponding night clip, is then fed into the same MLLM to produce diverse question candidates centered on the given QA type. 
This step encourages variety in phrasing and perspective while maintaining fidelity to the visual content.  

3) \textbf{Day-augmented pseudo answer synthesis.}  
For paired QA types, pseudo answers are generated by consulting the aligned daytime clip, where content is more visible and less ambiguous. For unpaired QA types or datasets without alignment (e.g., EgoNight-Oxford), answers are instead derived directly from the nighttime clip.  

All three stages are powered by GPT-4.1. 
Empirically, we find that both the QA-type-specific prompting and the inclusion of daytime videos substantially improve the quality and reliability of the generated QA pairs. More examples of VQA pairs and caption generation can be found in Appendix~\ref{supp:qa_examples} and ~\ref{supp:more_visualization}.

\noindent\textbf{Human Annotator Refinement. }
Finally, human annotators refine QA pairs via three operations: 
i) \textbf{delete}, when QA pairs are meaningless, vague, duplicated, or inconsistent across day–night counterparts (for paired QA types);  
ii) \textbf{modify}, when the question is valid but the answer is wrong (or vice versa), or to resolve ambiguity;
iii) \textbf{add}, when many pairs are removed or when important, challenging questions, especially about dynamic concepts, are missing.

After the first labeling round, we performed a random double-check to refine low-quality annotations. Thus, although our pipeline combines model generation with human refinement, \textit{every QA pair (3,658 in total) is manually verified at least once}. In total, $\sim$200 hours of human effort were invested, ensuring the quality and reliability of EgoNight-VQA.

\begin{figure}[htbp]
    \centering
    \includegraphics[width=\linewidth]{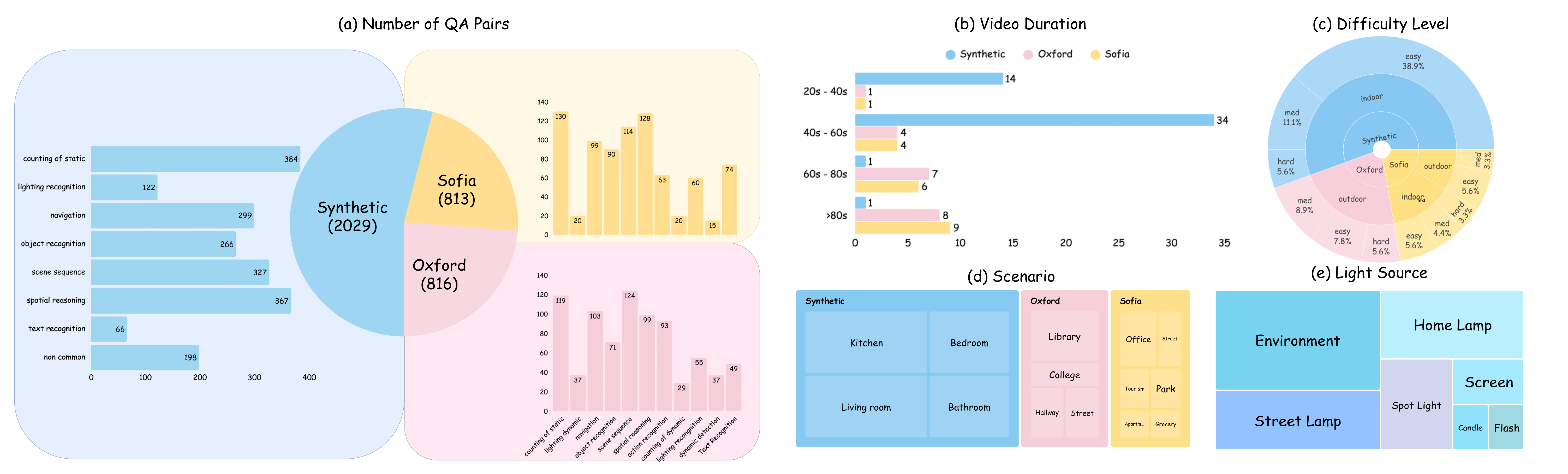}
    \caption{\textbf{Statistics of EgoNight-VQA benchmark.}  (a) Distribution of QA pairs across QA types and sources. (b) Video duration distribution. (c) Task difficulty levels cross scenarios. (d) Scenario coverage. (e) Illumination coverage.
    }
    \label{fig:statistics}
    \vspace{-0.1in}
\end{figure}

\noindent\textbf{Dataset Statistics. }
EgoNight-VQA comprises 3,658 high-quality, fully human-verified QA pairs across 12 task types, sourced from EgoNight-Synthetic, EgoNight-Sofia, and EgoNight-Oxford, with an average of 40 pairs per video. Detailed statistics on QA distribution, video durations, task difficulties, scenarios, and illumination are shown in Fig.~\ref{fig:statistics}. The number of videos across the three subsets—Synthetic (50), Sofia (20), and Oxford (20)—is proportionally reflected in the VQA annotations (2029 : 813 : 816). This results in an approximately 1:1 balance between synthetic and real (Sofia + Oxford) VQA samples, ensuring that our benchmark is not dominated by synthetic content. We provide more comparison of Egocentric VQA datasets in Appendix~\ref{sec:dataset_cmp}. Overall, EgoNight-VQA provides a diverse and comprehensive benchmark for evaluating egocentric vision models under nighttime conditions.

\subsection{Benchmarks Beyond Egocentric VQA}~\label{sec:beyond_vqa}
\noindent\textbf{Day-Night Correspondence Retrieval.}
To further assess model capabilities beyond VQA, we introduce \textit{day–night correspondence retrieval}, which evaluates a model’s ability to match visual content across illumination conditions. Specifically, we define two subtasks:
\textbf{i) Spatial Retrieval (Scene Recognition).} Spatial retrieval, or scene recognition, is a long-standing vision task~\cite{arandjelovic2016netvlad, scenegraphloc, DBLP}. Here, it is extended: given a query clip and a set of $N$ candidate clips of equal duration $s$, the model must retrieve the one depicting the same scene. This evaluates a model’s ability to capture and relate spatial relations in egocentric videos, e.g., distinguishing a bedroom from a bathroom or another bedroom. We built this benchmark with 1000 randomly generated meta-tasks. Each task samples a query clip, and the candidate set includes its temporally aligned counterpart (with a temporal shift for added difficulty) plus $N-1$ negatives from other scenes. Performance is measured by Top-1 accuracy across all tasks. In our setup, we use $N=10$, $s=10$ seconds, and a temporal shift of [10, 20] frames. Both \textit{Day (query) $\rightarrow$ Day (database)} and \textit{Day $\rightarrow$ Night} settings are evaluated.

\textbf{ii) Temporal Localization.}  
We further design a temporal localization task to test whether models can align video segments across dynamics. Given a query clip of duration $s$, the model must localize it within the corresponding full video by predicting its start and end timestamps $(t_i, t_j)$, directly evaluating temporal reasoning (e.g., grounding “The door is being locked” to 10–20s). We construct 1000 meta-tasks, each generated by randomly sampling one clip from its parent full video that is also randomly selected. Inspired by temporal grounding literature~\cite{st_v_groud}, we adopt mean Intersection-over-Union (mIoU) between the predicted interval $(t_i, t_j)$ and the ground-truth interval $(t_i^*, t_j^*)$ as the evaluation metric. Consistent with spatial retrieval, we set $s=10$ seconds and evaluate both  \textit{Day $\rightarrow$ Day} and \textit{Night $\rightarrow$ Day} settings.

\noindent\textbf{Egocentric Depth Estimation at Night.}
Depth estimation is a fundamental component of computer vision. On the one hand, extensive research~\cite{depth_anything_v1, depth_anything_v2, wang2025vggt} has focused on depth estimation in non-egocentric settings (typically not with fisheye cameras), while egocentric depth estimation remains largely underexplored, especially under nighttime conditions. On the other hand, recent works~\cite{chen2024spatialvlm, liu2025ssr} suggest that incorporating depth can enhance models’ spatial reasoning abilities. These two observations motivate us to construct an auxiliary benchmark for \textit{egocentric depth estimation at night}.  
Specifically, we use EgoNight-Synthetic as the testbed, where ground-truth depth maps are provided by the rendering engine. Thanks to the day–night aligned design, we can quantitatively evaluate models under both controlled daytime and nighttime conditions. For evaluation, we adopt standard depth estimation metrics, including absolute relative error (AbsRel), $\delta_1 (1.25)$, $\delta_2 (1.25^2)$, and $\delta_3 (1.25^3)$, where $\delta_k$ measures the percentage of predicted pixels whose relative error is within a threshold of $1.25^k$.

\vspace{-1em}

\section{Experiments}

\subsection{Evaluated MLLMs \& Metrics}~\label{sec:setup}
We evaluate a broad set of state-of-the-art MLLMs on the proposed benchmarks.
i) For \textbf{EgoNight-VQA}, we include two closed-source commercial models, GPT-4.1~\cite{achiam2023gpt} and Gemini 2.5 Pro~\cite{comanici2025gemini}; eight open-source models, Qwen2.5-VL (3B, 7B, 72B)~\cite{Qwen-VL}, VideoLLaMA3 (7B)~\cite{zhang2023video}, InternVL3 (8B)~\cite{chen2024internvl}, GLM-4.1V (9B-Base)~\cite{hong2025glm}, and LLaVA-NeXT-Video (7B)~\cite{li2024llava}; as well as EgoGPT~\cite{yang2025egolife}, one of the few open-source egocentric models tailored for open-ended QA. Following prior work~\cite{plizzari2025omnia,fan2019egovqa}, we adopt an \textit{LLM-as-a-Judge} strategy to assess semantic consistency between predictions and ground truth, and report average accuracy across the test sets. ii)We provide further in-depth analysis on synthetic data quality, and potential model improvements. iii) For \textbf{day–night correspondence retrieval}, we benchmark feaFVisture-based retrieval methods, DINOv2~\cite{oquab2023dinov2} and Perception Encoder (Percep. Enc.)~\cite{bolya2025perceptionencoderbestvisual}, alongside MLLM-based methods, GPT-4.1 and InternVL3 (8B). As described in Sec.~\ref{sec:beyond_vqa}, Top-1 accuracy (Acc-R@1, \%) and mIoU (\%) are used for evaluating the spatial and temporal subtasks, respectively.
iv) For \textbf{egocentric depth estimation}, we test a general monocular depth model (Depth Anything~\cite{depth_anything_v1,depth_anything_v2}), a 3D reconstruction-based method (VGGTStream~\cite{streamVGGT,wang2025vggt}), and two egocentric fisheye-specific models (DAC~\cite{Guo2025DepthAnyCamera} and UniK3D~\cite{piccinelli2025unik3d}). For Depth Anything and VGGTStream, input fisheye RGB frames and depth maps are undistorted prior to inference for fair comparison. Additional implementation details (e.g., fps for frame extraction, prompts, and model settings) and discussion about LLM-as-a-Judge strategy are provided in the Appendix Sec.~\ref{sec:more_experiment_setup}.

\begin{table*}[t] 
    \centering
    \scalebox{0.9}{
    \begin{tabular}{l|ccc|ccc|ccc|c}
        \toprule
        \multirow{2}{*}{\textbf{Models}} 
        & \multicolumn{3}{c|}{{\textbf{EgoNight-Synthetic}}} 
        & \multicolumn{3}{c|}{{\textbf{EgoNight-Sofia}}} 
        & \multicolumn{3}{c|}{{\textbf{EgoNight-Oxford}}} 
        & \multicolumn{1}{c}{{\textbf{Avg.}}} \\
        & Easy & Medium & Hard & Easy & Medium & Hard & Easy & Medium & Hard & - \\ 
        \midrule
\multicolumn{11}{c}{\textit{Closed-Source MLLMs}} \\
\midrule
\multicolumn{1}{l|}{GPT-4.1} & 29.30 & \cellcolor{blue!10} \textbf{26.87} & \cellcolor{blue!10} \textbf{18.87} & 32.04 & \cellcolor{blue!10} \textbf{29.35} & \cellcolor{blue!10} \textbf{31.69} & \cellcolor{blue!10} \textbf{39.72} & \cellcolor{blue!10} \textbf{37.13} & \cellcolor{blue!10} \textbf{40.72} & \cellcolor{blue!10} \textbf{30.93} \\
\multicolumn{1}{l|}{Gemini 2.5 Pro}  & \cellcolor{blue!10} \textbf{31.05} & 24.81 & 16.51 & \cellcolor{blue!10} \textbf{38.24} & 26.81 & 28.87 & 36.75 & 36.81 & 27.88 & 30.60 \\
\midrule
\multicolumn{11}{c}{\textit{Open-source MLLMs}} \\
\midrule
\multicolumn{1}{l|}{InternVL3-8B} & \cellcolor{blue!10} 20.21 & \cellcolor{blue!10} 15.50 & \cellcolor{blue!10} 16.98 & \cellcolor{blue!10} 24.03 & \cellcolor{blue!10} 21.74 & \cellcolor{blue!10} 20.42 & 22.90 & 20.85 & 16.36 & \cellcolor{blue!10} 20.06 \\

\multicolumn{1}{l|}{Qwen2.5-VL-72B} & 18.39 & 15.25 & 12.26 & \cellcolor{blue!10} 24.03 & 17.03 & \cellcolor{blue!10} 20.42 & \cellcolor{blue!10} 24.81 & \cellcolor{blue!10} 22.80 & 16.36 & 18.99 \\

\multicolumn{1}{l|}{Qwen2.5-VL-7B} & 13.01 & 13.95 & 13.68 & 15.44 & 12.68 & 12.68 & 13.74 & 13.36 & 12.73 & 13.44 \\

\multicolumn{1}{l|}{Qwen2.5-VL-3B} & 14.69 & 10.34 &  7.08 & 15.50 & 13.04 & 12.68 & 17.18 & 11.40 & 12.12 & 13.41 \\

\multicolumn{1}{l|}{GLM-4.1V-9B-Base} & 19.09 & 13.70 & 15.57 & 18.60 & 18.48 & 16.20 & 17.15 & 22.15 & \cellcolor{blue!10} 18.79 & 18.20 \\

\multicolumn{1}{l|}{VideoLLaMA3-7B} & 16.85 & 13.44 & 14.62 & 11.11 & 10.87 &  9.15 & 12.26 & 10.46 &  9.15 & 13.64 \\
\multicolumn{1}{l|}{LLaVA-NeXT-Video-7B} & 6.36 & 11.37 &  1.89 & 13.95 &  9.78 & 14.79 &  3.05 &  2.61 &  3.03 &  7.28 \\

\midrule
\multicolumn{11}{c}{\textit{Egocentric MLLMs}} \\
\midrule
\multicolumn{1}{l|}{EgoGPT} & \cellcolor{blue!10}15.79 &\cellcolor{blue!10} 13.55 & \cellcolor{blue!10}12.04 & \cellcolor{blue!10}12.41 & \cellcolor{blue!10}12.13 & \cellcolor{blue!10}10.36 & \cellcolor{blue!10}12.37 & \cellcolor{blue!10}13.58 & \cellcolor{blue!10}13.68 & \cellcolor{blue!10}14.29 \\
\bottomrule
\end{tabular}
}
\vspace{-0.05in}
\caption{\textbf{Comparison results on EgoNight-VQA.} Accuracies (\%) of OpenQA results across three datasets and three difficulty levels. We compare closed-source models, open-source models, and egocentric-specific models.
} \label{tab:main_result}
\vspace{-0.15in}
\end{table*}

\vspace{-1em}
\subsection{Results on EgoNight-VQA}~\label{sec:vqa_result}
\begin{figure}[h]
    \centering
    \includegraphics[width=0.9\linewidth]{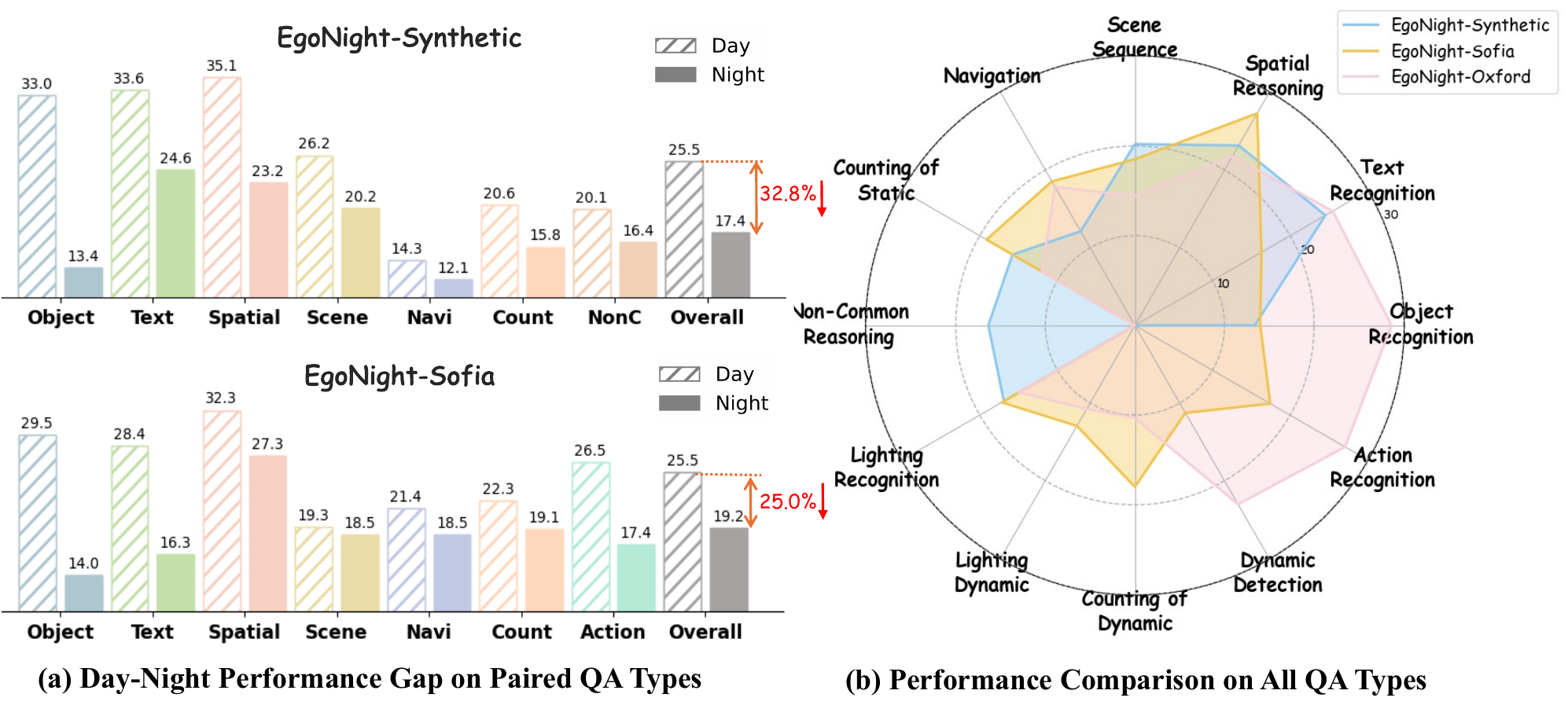}
    \vspace{-0.15in}
     \caption{
    {\textbf{Performance analysis of MLLMs on EgoNight-VQA.} (a) Day–night performance gap across paired QA types, showing consistent degradation at night. (b) Nighttime performance across all 12 QA types. NonC means non–common-sense reasoning.}}
    
    \label{fig:gap}
    \vspace{-0.1in}
\end{figure}

The main results of all MLLMs are shown in Tab.~\ref{tab:main_result}. In addition, we provide per-QA performance comparisons between day (striped bars) and night (solid bars) for paired QA types (Fig.~\ref{fig:gap}(a)) and report nighttime performance across all QA types (Fig.~\ref{fig:gap}(b)), based on averages across all models. Note that non–common case detection is available only in EgoNight-Synthetic, while dynamic events and actions are included only in the real-world data.

From the results in Tab.~\ref{tab:main_result}, we observe that almost all MLLMs struggle on our benchmark, with maximum averaged accuracies of 30.93\% from the closed-source GPT-4.1, 20.06\% from the open-source InternVL3-8B, and 14.29\% from the egocentric EgoGPT. The wide performance spread also confirms that our dataset is sufficiently challenging and effective for distinguishing model capabilities. Fig.~\ref{fig:gap}(a) further highlights the performance gap, showing declines of 32.8\% and 25.0\% on EgoNight-Synthetic and EgoNight-Sofia, respectively. Together, these results underscore the substantial challenges posed by our benchmark, exposing the limitations of current MLLMs under nighttime scenarios and highlighting the need for more illumination-robust models. Beyond the overall trends, we note three additional insights from Tab.~\ref{tab:main_result}: i) Closed-source models perform best. Within open-source models, Qwen2.5-VL generally improves with scale, yet InternVL outperforms the larger Qwen2.5-VL (72B), suggesting that size alone is insufficient. The relatively low results of EgoGPT further emphasize the need for more robust egocentric models. ii) EgoNight-Oxford achieves the highest scores, but its illumination conditions are more challenging than those in EgoNight-Synthetic and EgoNight-Sofia (Sec.~\ref{supp:qa_examples}, Appendix). This indicates that without paired day videos and our day-augmented auto-labeling strategy, even human annotators face difficulties generating challenging QA pairs, underscoring the practical value of our dataset design; iii) Overall, performance declines across task levels (easy, medium, hard), validating the diversity and difficulty of our benchmark.

From the per-QA results in Fig.~\ref{fig:gap}(a) and Fig.~\ref{fig:gap}(b), we further observe three key trends:
i) Models perform better on perception-oriented tasks (e.g., object recognition, text recognition, scene sequence) than reasoning-oriented tasks (e.g., navigation, counting, non-common-sense reasoning cases) under daytime conditions. However, at night, perception tasks suffer larger performance drops, indicating their higher sensitivity to illumination, whereas reasoning tasks, though harder overall, are relatively less affected since they rely more on temporal and contextual cues.
ii) MLLMs achieve substantially lower accuracy on our newly proposed tasks, such as lighting recognition, lighting dynamics, scene sequence, dynamic detection, navigation, and non-common-sense reasoning, suggesting that existing MLLMs generalize poorly to novel tasks compared with well-studied ones like object recognition.
iii) Each dataset in Fig.~\ref{fig:gap}(b) emphasizes distinct aspects of nighttime challenges, together providing complementary perspectives that ensure EgoNight spans a balanced range of perception–reasoning difficulties under low-light conditions.

\subsection{More in-depth analysis}~\label{sec:indepth_analysis}

\noindent\textbf{Quality of EgoNight-Synthetic.} More examples of synthetic visualizations are provided in Appendix~\ref{supp:qa_examples} and~\ref{sec:more_construction}. To quantitatively evaluate how well the synthetic data reflects real-world performance, we benchmark a diverse set of multimodal large language models (MLLMs) on three subsets: Synthetic, Sofia (real), and Oxford (real). For each model, we compute its overall VQA accuracy on each subset and then calculate the Pearson correlation between the average per-model accuracies on synthetic and real-world data (Appendix~\ref{sec:more_exp_result}). We observe strong correlations between synthetic and Sofia ($0.9359$, p-value $6.847\times10^{-5}$), and between synthetic and Oxford ($0.8588$, p-value $1.462\times10^{-3}$). These results indicate that models that perform well on synthetic data also tend to perform well on real nighttime data, suggesting that the synthetic set preserves the relative difficulty and ranking of models.

To further examine transferability, we fine-tune Qwen2.5-VL-7B using supervised fine-tuning (SFT) on the synthetic training split only, while keeping the real-world data unseen during training. We then evaluate the fine-tuned model directly on the real dataset. The accuracy improves from $14.83\%$ (zero-shot baseline) to $20.57\%$, demonstrating that supervision from synthetic nighttime scenes generalizes to real-world nighttime scenarios.

\noindent\textbf{Impact of Fine-Tuning.} 
To study model adaptation to nighttime egocentric QA, we split EgoNight into $70\%$ training and $30\%$ testing subsets, ensuring no video overlap between splits. We fine-tune Qwen2.5-VL-7B with supervised fine-tuning (SFT) under three configurations: i) full model fine-tuning (updating both vision encoder and LLM), ii) vision-encoder-only tuning (freezing the LLM), and iii) LLM-only tuning (freezing the vision encoder). All models are trained on the same training split and evaluated on the held-out test split.

Tab.~\ref{tab:finetune_all} reports the performance for synthetic and real (Sofia and Oxford) separately. It shows consistent performance gains across all configurations compared to the zero-shot baseline. Updating either the vision encoder or the LLM independently improves accuracy, while full fine-tuning achieves the best performance, indicating that both improved visual representations and stronger vision-language alignment contribute to adaptation under low-light conditions. 

\noindent\textbf{Perception vs.\ Reasoning under Fine-Tuning.} 
To better understand which capabilities benefit most from adaptation, we group tasks into perception-oriented (Object Recognition, Text Recognition) and reasoning-oriented (Navigation, Counting) categories and report their accuracies separately in Tab.~\ref{tab:finetune_task}. 

Perception-oriented tasks show larger absolute improvements after fine-tuning, suggesting that low-level visual adaptation plays a significant role in nighttime scenarios. Vision-encoder-only tuning primarily boosts perception tasks, whereas LLM-only tuning improves both perception and reasoning tasks to a certain extent. However, reasoning-oriented tasks exhibit comparatively smaller gains overall, indicating that high-level reasoning under degraded visual input remains challenging even after adaptation.

\noindent\textbf{Failure Cases.} Additional analysis of failure cases is provided in Appendix~\ref{supp:failure_case}.

\begin{table}[ht]
\centering
\begin{minipage}{0.4\textwidth}
\centering
\begin{tabular}{lcc}
\hline
\textbf{Model} & \textbf{Synthetic} & \textbf{Real} \\
\hline
Qwen7B (Base) & 23.23 & 16.40 \\
Enc. Only & 29.74 & 20.92 \\
LLM. Only & 35.50 & 22.26 \\
Full & \cellcolor{blue!10}{36.25} & \cellcolor{blue!10}{25.61} \\
\hline
\end{tabular}
\caption{Fine-tuning performance comparison across datasets.}
\label{tab:finetune_all}
\end{minipage}
\hspace{0.05\textwidth}
\begin{minipage}{0.4\textwidth}
\centering
\begin{tabular}{lccc}
\hline
\textbf{Task} & \textbf{Qwen7B} & \textbf{Enc. Only} & \textbf{LLM. Only} \\
\hline
Object     & 8.435  & 34.718 & \cellcolor{blue!10}{35.855} \\
Text       & 18.440 & 49.890 & \cellcolor{blue!10}{50.988} \\
Navigation & 17.870 & 19.495 & \cellcolor{blue!10}{19.918} \\
Counting   & 16.558 & 16.945 & \cellcolor{blue!10}{24.275} \\
\hline
\end{tabular}
\caption{Fine-tuning performance comparison across tasks.}
\label{tab:finetune_task}
\end{minipage}
\end{table}

\subsection{Results on Day-Night Correspondence Retrieval}~\label{sec:retrieval_result}

\begin{table}[ht]
\vspace{-0.1in}
\centering
\small
\scalebox{.9}{
\begin{tabular}{l|cc|cc|cc|cc}
\toprule
\multirow{2}{*}{\textbf{Models}} 
& \multicolumn{4}{c|}{\textbf{Spatial Retrieval (Acc - R@1 $\%$ ↑)}} 
& \multicolumn{4}{c}{\textbf{Temporal Localization (mIoU $\%$ ↑)}} \\
\cmidrule(lr){2-5} \cmidrule(lr){6-9}
& \multicolumn{2}{c}{EgoNight-Synthetic} 
& \multicolumn{2}{c|}{EgoNight-Sofia} 
& \multicolumn{2}{c}{EgoNight-Synthetic} 
& \multicolumn{2}{c}{EgoNight-Sofia} \\
\cmidrule(lr){2-3} \cmidrule(lr){4-5} \cmidrule(lr){6-7} \cmidrule(lr){8-9}
& Day$\to$Day & Night$\to$Day 
& Day$\to$Day & Night$\to$Day 
& Day$\to$Day & Night$\to$Day 
& Day$\to$Day & Night$\to$Day \\
\midrule
DINOv2 & 45.7 & 28.7 & 84.5 & 74.5  & - & \cellcolor{blue!10}33.7 & - & 33.1 \\
Percep. Enc. & 65.4 & 41.6 & 89.8 & 80.9 & - & 32.9 & - & \cellcolor{blue!10}33.4  \\
GPT-4.1         & \cellcolor{blue!10}75.6   & \cellcolor{blue!10}54.1  & \cellcolor{blue!10}92.5  & \cellcolor{blue!10}84.5 & \cellcolor{blue!10}14.7  & 10.0 & \cellcolor{blue!10}21.2 & 15.5\\
InternVL3-8B    &   39.4   & 27.7 & 73.9 & 56.3 & 10.2 & 9.9 & 12.5 & 13.3 \\
\bottomrule
\end{tabular}}
\vspace{-0.05in}
\caption{\textbf{Night-to-Day retrieval performance.} 
Each dataset is evaluated on both Day$\to$Day and Night$\to$Day settings.}
\label{tab:retrieval}
\end{table}

The results of day–night retrieval are reported in Tab.~\ref{tab:retrieval}. The gap between Night–Day and Day–Day shows that cross-illumination retrieval remains highly challenging compared with in-domain retrieval. For spatial retrieval, GPT-4.1 consistently outperforms other methods, achieving over $80\%$ accuracy. This suggests that Retrieval-Augmented Generation methods could further improve performance, as Fig.~\ref{fig:gap}(a) already shows that daytime inputs significantly benefit the models.  For temporal retrieval, however, GPT-4.1, despite its strong results on egocentric VQA (Tab.~\ref{tab:main_result}) and spatial retrieval, shows a substantial drop compared with feature-based methods (DINOv2 and Perception Encoder). A similar degradation is observed for InternVL3-8B. These findings suggest that while MLLMs excel at spatial semantic understanding, they struggle with temporal reasoning, such as timestamp prediction, which is critical for temporal localization. Further results on temporal limitations are provided in Appendix~\ref{sec:more_exp_result}.

\subsection{Results on Depth Estimation}~\label{sec:depth_result}
Results for depth estimation are reported in Tab.~\ref{tab:egonight_depth}. The relatively low scores across all models highlight the difficulty of our EgoNight dataset, which combines egocentric motion, complex geometry, and extreme lighting variations. A clear gap between daytime and nighttime performance again underscores the challenges of low-light conditions. Among the methods, fisheye-based methods (DAC and UniK3D) outperform general depth estimators, suggesting the need for egocentric-specific algorithms. Additional qualitative results are provided in Sec.~\ref{sec:EgoDepthQualitative}.

\begin{table}[ht]
\centering
\vspace{-0.1in}
\scalebox{.8}{
\begin{tabular}{l|cc|cc|cc|cc}
\toprule
\multirow{2}{*}{\textbf{Method}} 
& \multicolumn{2}{c|}{\textbf{Abs Rel ↓}} 
& \multicolumn{2}{c|}{\textbf{$\delta_1$} (1.25) ↑} 
& \multicolumn{2}{c|}{$\delta_2$ (1.25$^2$) ↑} 
& \multicolumn{2}{c}{$\delta_3$ (1.25$^3$) ↑} \\
\cmidrule(lr){2-3} \cmidrule(lr){4-5} \cmidrule(lr){6-7} \cmidrule(lr){8-9}
& Day & Night & Day & Night & Day & Night & Day & Night \\
\midrule
Depth Anything  \textcolor{blue}{(\textbf{U})}
& 0.297 & 0.302 & 0.249 & 0.237 & 0.463 & 0.447 & 0.622 & 0.60\ \\
VGGTStream \textcolor{blue}{(\textbf{U})}
& 0.293 & 0.298 & 0.234 & 0.232 & 0.447 & 0.442 & 0.615 & 0.609 \\
DAC \textcolor{green!60!black}{(\textbf{F})}
& 0.245 & 0.292 & 0.255 & 0.216 & 0.495 & 0.425 & 0.684 & 0.602 \\
UniK3D \textcolor{green!60!black}{(\textbf{F})} 
& \cellcolor{blue!10}0.224 & \cellcolor{blue!10}0.253 & \cellcolor{blue!10}0.280 &\cellcolor{blue!10} 0.254 & \cellcolor{blue!10}0.524 & \cellcolor{blue!10}0.481 & \cellcolor{blue!10}0.706 & \cellcolor{blue!10}0.658 \\
\bottomrule
\end{tabular}}
\vspace{-0.05in}
\caption{
Depth estimation results on EgoNight-Synthetic. \textcolor{blue}{\textbf{U}}: undistorted input; \textcolor{green!60!black}{\textbf{F}}: fisheye input. }
\label{tab:egonight_depth}
\vspace{-0.15in}
\end{table}

\section{Conclusion}
\vspace{-1em}
In this work, we introduced EgoNight, the first benchmark suite designed to systematically evaluate egocentric multimodal large language models (MLLMs) under challenging nighttime conditions. EgoNight integrates synthetic and real-world videos with day–night alignment, enabling rigorous analysis of illumination effects. Building upon this data, we proposed EgoNight-VQA, spanning 12 QA types with 3,658 human-verified pairs, alongside two complementary benchmarks: day–night correspondence retrieval and egocentric depth estimation. Experiments reveal that even state-of-the-art MLLMs struggle under low-light conditions, with performance dropping substantially compared to daytime. This highlights that nighttime egocentric vision remains far from being solved, motivating future research into illumination-robust egocentric perception and reasoning. We believe EgoNight provides a valuable and timely benchmark that will drive progress toward more reliable egocentric AI assistants.

\subsubsection*{Acknowledgments}
This research was partially funded by a collaboration project between VIVO and INSAIT on indoor scene understanding and editing. Research was also funded by the Ministry of Education and Science of Bulgaria (support for INSAIT, part of the Bulgarian National Roadmap for Research Infrastructure) . This project was supported with computational resources provided by Google Cloud Platform (GCP). We sincerely thank Dr. Zirui Wang, the author of the Oxford Day-and-Night dataset, for the helpful discussions and for kindly providing the source videos.

\bibliography{ref}
\bibliographystyle{iclr2026_conference}

\appendix

\title{xxx}

\section{Appendix}
\subsection{More Video Source Construction Details}
\label{sec:more_construction}
\noindent\textbf{EgoNight-Synthetic Construction.}
For EgoNight-Synthetic Construction, we first use the coarse progressive generation method with a fast solver in infinigen~\cite{infinigen2023infinite} to generate 3D scenes in Blender format. Then, a human annotator will edit the scene in the following sequence:
\begin{itemize}
    \item Explore and edit the scene to remove unreasonable cases and make the indoor scene as natural as possible. 
    \item Add light source in the scene if the generated scene does not include enough illumination to create enough illumination gap between the day and night. 
    \item Record camera trajectory by exploring the whole indoor scene. 
    \item Change the camera model and resolution. Set rendering samples and frames. For all synthetic dataset, we use the Blender build-in Panoramic Fisheye Equisolid camera with Lens $10.5$ and field of view $180$\textdegree.
    \item Create night scene by modifying the light source, motion blur, and environment map.
    \item Render the day and night pair using Blender~\cite{iraci2013blender}. 
\end{itemize}

\begin{figure}[h]
    \centering
    \includegraphics[width=0.9\linewidth]{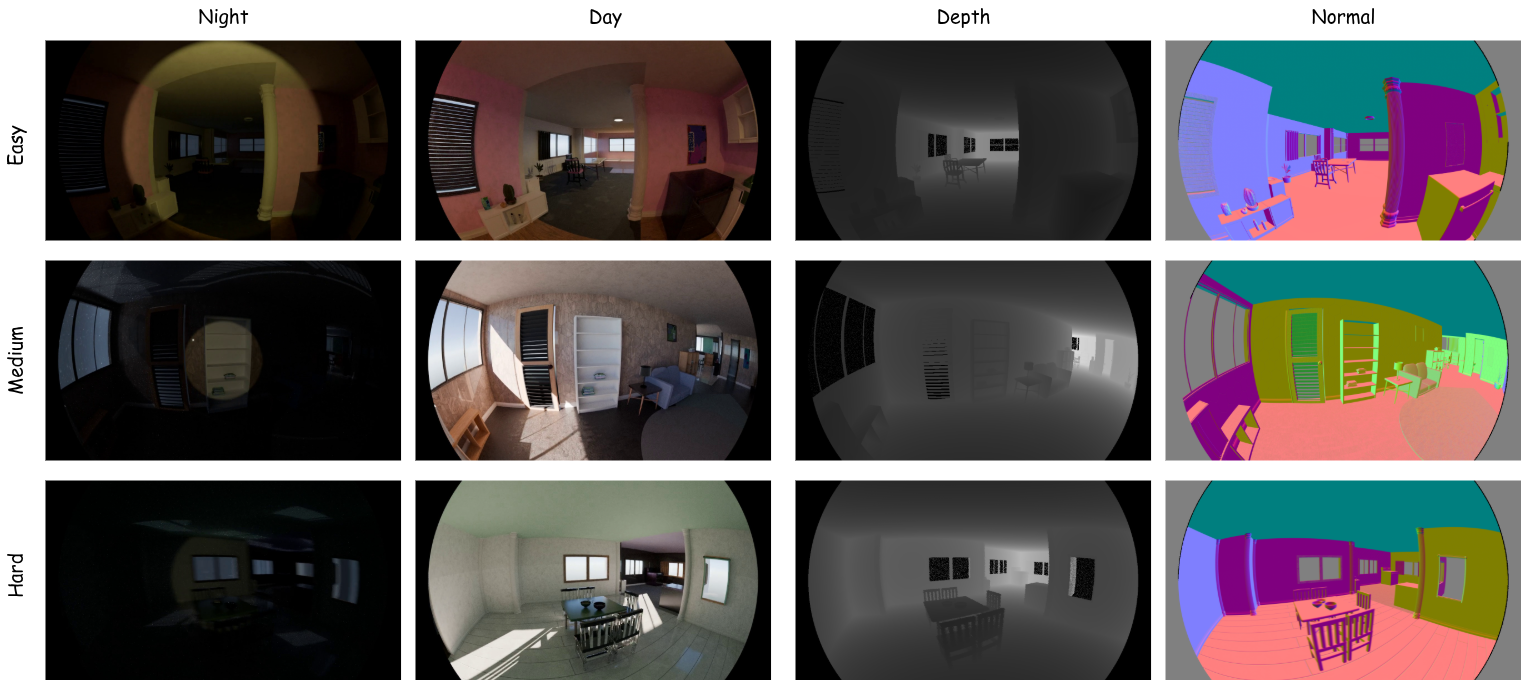}
    \vspace{-0.05in}
    \caption{
     More examples and modalities of synthetic datasets.
    }
    \label{fig:more_synthetic}
    \vspace{-0.1in}
\end{figure}

During the dataset construction, we apply home light source during night for 30 scenes and spot light source for 20 scenes to simulate torch light in real life. To create different difficulty levels, we apply different rendering sample size (higher sample size gives lower noise in the final image), spot light size, and motion blur to part of the data as shown in Tab.~\ref{tab:synthetic_difficulty_levels}. We also show different modality and difficulty level in Fig.~\ref{fig:more_synthetic}

\begin{table}[h!]
\centering\small
\vspace{-0.1in}
\begin{tabular}{c|c|c|c}
\hline
 \textbf{difficulty level} &\textbf{sample size} & \textbf{light condition} & \textbf{motion blur shutter} \\ \hline
Easy   & 4096 & 105\textdegree / few light on & - \\ \hline
Medium & 512 & 40\textdegree-50\textdegree spot light / all light off& - \\ \hline
Hard   & 512 &40\textdegree-50\textdegree spot light / all light off & 1-2\\ \hline
\end{tabular}
\caption{Difficulty level and corresponding rendering settings.}
    \vspace{-0.1in}
\label{tab:synthetic_difficulty_levels}
\end{table}

\noindent\textbf{EgoNight-Sofia Construction}. 
In total, four participants were involved. The recording setup included three different GoPros, a head-mounted rig to fix the camera on the forehead and mimic human-eye perspective, several phones for live preview or daytime video guidance, and diverse lighting sources such as flashlights, spotlights, and candles. The process followed a video-guided recording strategy, as introduced in ~Sec.~\ref{sec:video_source}: the ego-wearer first recorded a daytime video while previewing the live feed on a phone, and for the nighttime counterpart replayed the daytime video on the phone as guidance to replicate the same setup, walking speed, viewpoints, and actions. Videos were collected across a wide range of environments, including indoor scenes (apartments, workplaces, grocery shops, building receptions) and outdoor scenarios (fitness areas, tourist landmarks, and street views). Post-trimming was applied to each day–night pair to further ensure alignment. On average, it took around 2-3 hours to produce one paired data.

\noindent\textbf{EgoNight-Oxford Construction.}
We credit the contribution of Seeing in the Dark dataset~\cite{wang2025seeing}, which provides multiple sequences of egocentric videos in the night in a various environment. We built our EgoNight-VQA dataset partially upon this work. Firstly, We enumerated all nighttime clips in Oxford Day–Night and performed a two-stage filtering. (1) Screening for uniqueness of place. We cross-checked scene metadata (route notes/time stamps) to avoid repeated paths within the same landmark. (2) Stratified diversity \& quality sampling. Remaining clips were scored on a 1–5 rubric along axes designed for egocentric, low-light evaluation: illumination type (ambient only / mixed artificial / high-contrast point sources), illumination hardness (soft vs. specular/point), exposure stability (auto-gain pumping, blown highlights), scene dynamics (pedestrians/vehicles, occlusions), camera motion pattern (walk, run, head turns), and task context (navigation, road-crossing, object interaction, signage reading). The final set comprises 20 sequences that maximize lighting/task diversity under egocentric night settings while avoiding place overlap and task overlap.

In total, collecting the three video sources required over 100 hours of human effort.

\subsection{More Benchmark Implementation Details}

\begin{figure}[h]
    \vspace{-0.1in}
    \centering
    \includegraphics[width=1.\linewidth]{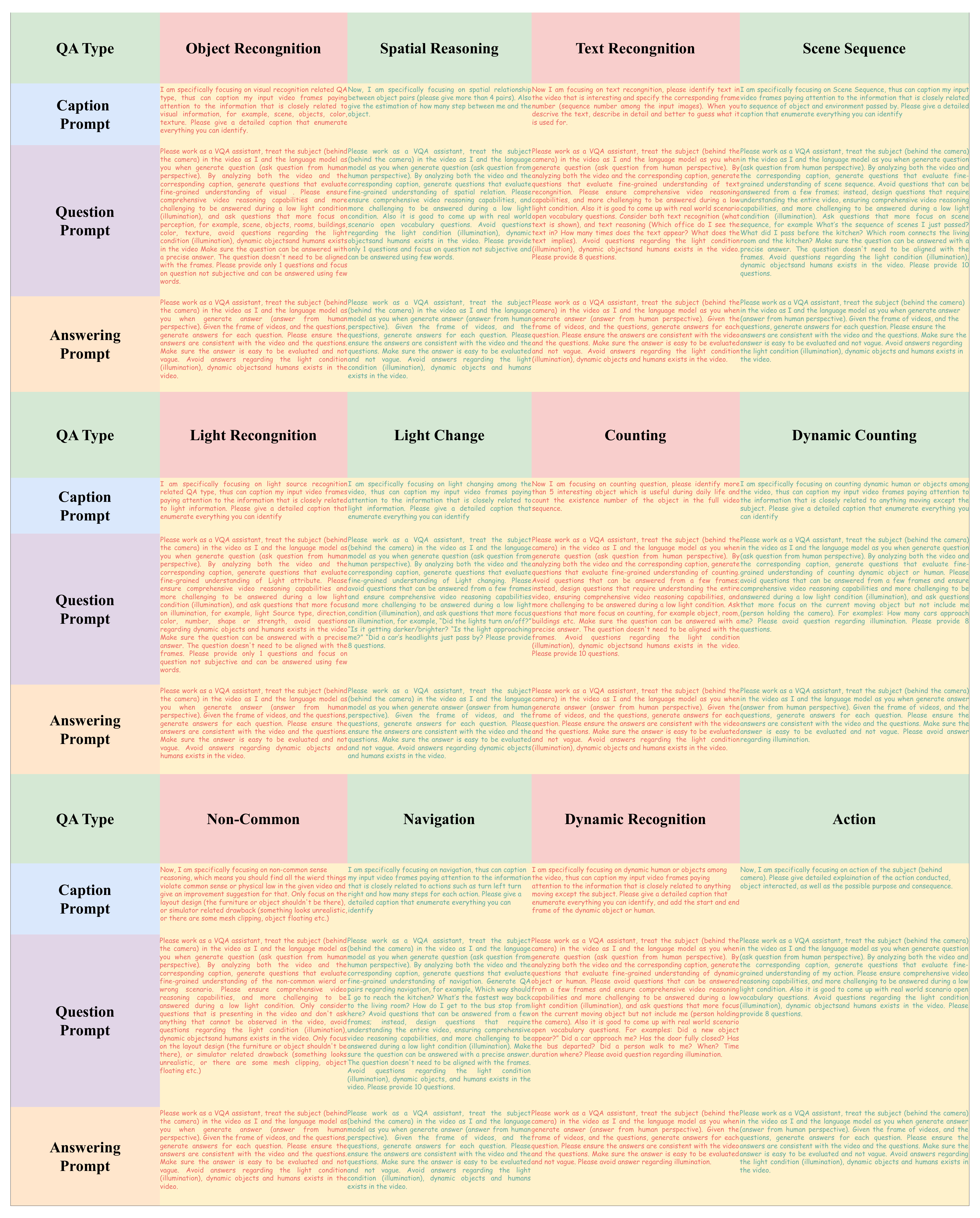}
     \vspace{-0.25in}
    \caption{
     The prompts used during auto-labeling.
    }
    \label{fig:labeling_prompt}
    \vspace{-0.3in}
\end{figure}

As described in Sec.~\ref{sec:vqa}, our auto-labeling pipeline is QA-type specific and involves three customized prompts: one for captioning, one for question generation, and one for answer synthesis. The detailed prompts are shown in Fig.~\ref{fig:labeling_prompt}.

\subsubsection{EgoNight-VQA Human Labeling}
We hired several participants to review and refine the QA pairs generated by our three-stage day-augmented auto-labeling pipeline, compensating them at a rate of €20 per video. Each participant was provided with a detailed labeling instruction document and an onboarding meeting to ensure the guidelines were clearly conveyed. A simplified version of the labeling tutorial is included as in Fig.~\ref{fig:tu}.

\begin{figure*}[h!]
\small\centering
\begin{tcolorbox}[width=1.\textwidth, colback=gray!5,colframe=black!40,title=Annotation Tutorial (Simplified)]
\textbf{Read Me First:} Please follow the labeling pipeline carefully and complete each step as instructed.
On average, annotating one video takes about 2 hours. Easier cases may take less time, but in general, each video should take no less than 1.5 hours to ensure high-quality annotations. (The first video may take longer, as you will need to familiarize yourself with the pipeline.)
We will randomly check the labeled data afterward, and annotators will be required to refine their work if the quality does not meet expectations.

\noindent\textbf{Step 1: Preparation.}  
You are expected to first download the paired \texttt{day.mp4} and \texttt{night.mp4} videos 
(aligned in time, except for unpaired tasks), together with the QA text file (\texttt{.txt}), 
which contains candidate QAs grouped by QA type (e.g., \texttt{counting.txt}).  
Before starting annotation, you should carefully watch both the daytime and nighttime videos 
to fully understand the scenario and activities.

\noindent\textbf{Step 2: QA Verification and Refinement.}  
For each QA pair, you should apply one of three operations:
\begin{itemize}
  \item \textbf{Delete}: You should remove QAs that are meaningless, vague, irrelevant, 
  duplicated, or inconsistent between day–night pairs (for paired QA types).  
  \item \textbf{Modify}: If the question is reasonable but the answer is incorrect 
  (e.g., counting errors, wrong action duration), you should correct the answer.  
  You may also rephrase the question to eliminate ambiguity 
  (e.g., clarifying ``left/right'' as relative to the ego-wearer).  
  \item \textbf{Add}: If too many pairs are deleted, or if you notice interesting and 
  challenging cases missing, you need to add new QAs.  
  This is especially important for low-frequency tasks, e.g., dynamic detection 
  or counting of dynamics.
\end{itemize}

\noindent\textbf{Step 3: Special Cases.}  
\begin{itemize}
  \item For paired QA types (e.g., object recognition, spatial reasoning), 
  you must ensure the same QAs apply to both day and night videos.  
  \item For unpaired QA types (e.g., lighting recognition, dynamic detection), 
  you only need to ensure correctness on the nighttime video.  
  \item For dynamic events, you are expected to specify temporal spans, e.g.,  
  \textit{Q: ``Around which time does a red car pass by?'' A: ``At frames 4--6.''}
\end{itemize}

\noindent\textbf{Step 4: Post-processing.}  
Once QAs were validated, the answer field should be renamed from \texttt{``answer''} 
to \texttt{``human\_answer''}.  

\noindent\textbf{Appendix: Paired \& Non-Paired QA Types.}
The same as described in the main file.
\end{tcolorbox}
\vspace{-0.1in}
~\caption{Simplified version of annotation tutorial. } ~\label{fig:tu}
\vspace{-0.2in}
\end{figure*}

After the first round of labeling, we performed an additional quality check and summarized common issues for refinement. For example, the concept of “left” or “right” should always be defined relative to the ego-wearer, and questions should be phrased clearly to avoid ambiguity. Participants were then asked to address the identified issues. In the end, we ensured that every QA pair in the EgoNight-VQA dataset was verified by at least one human annotator. On average, above 200 hours of human annotation effort were spent for the labeling refinement. To further evaluate inter-annotator statistics, we select randomly VQA pairs from 20 videos out of 90, and add four human annotators without GPT annotation. The average pairwise cosine similarity normalized to $(0,1)$ based on language feature extracted by BLIP-2~\cite{li2023blip2bootstrappinglanguageimagepretraining} is 0.8458. This shows that human annotators are in general consistent in interpreting scenes. And answers are semantically aligned, even when worded differently.

\subsection{Comparisons with Prior Egocentric VQA Benchmarks.}~\label{sec:dataset_cmp}
In Tab.~\ref{tab:benchmark_comparison}, we compare our EgoNight-VQA with prior egocentric VQA benchmarks, including  EgoVQA~\cite{fan2019egovqa}, EgoTaskQA~\cite{jia2022egotaskqa}, EgoSchema~\cite{mangalam2023egoschema}, EgoThink~\cite{cheng2024egothink}, EgoTempo~\cite{plizzari2025omnia},
EgoCross~\cite{li2025egocross}, EgoMemoria~\cite{ye2024mm}, HourVideo~\cite{chandrasegaran2024hourvideo}, and EgoLifeQA~\cite{yang2025egolife}, listing their lighting conditions (mainly daytime or nighttime), video duration length, the number of testing QA examples, number of QA type categories, if temporal-oriented tasks are included or emphasized, and the evaluation metric.

We highlight that EgoNight-VQA is the first to explore nighttime egocentric VQA with aligned day–night video pairs. 

\begin{table*}[htbp]
\small
\scalebox{.95}{
\centering\small
\begin{tabular}{lcccccc}
\toprule
\textbf{Dataset} & \textbf{Lighting} & \textbf{Video Length} & \textbf{\# Test} & \textbf{\# Categories} & \textbf{Temporal} & \textbf{Metric Type}\\
\midrule
EgoVQA     & \faSunO   & (25s, 100s) & 250  & 3  & \ding{55} & OpenQA    \\
EgoTaskQA  & \faSunO   & 25s         & 8k   & 4  & \ding{55} & OpenQA    \\
EgoSchema  & \faSunO   & 180s       & 500  & -  & \ding{55} & CloseQA    \\
EgoThink  & \faSunO   & -           & 750  & 12 & \ding{55} & OpenQA    \\
EgoTempo  & \faSunO  & 45s         & 500  & 10 & \ding{51} & OpenQA      \\
EgoCross & \faSunO  & 23s       & 957  & 15 & \ding{51} & CloseQA \& OpenQA \\
EgoBlind& \faSunO   & (0s, 120s) & 5311 & 6 & \ding{51} & OpenQA \\
EgoMemoria & \faSunO  & (30s, 1h) & 7026 & - & \ding{51} & CloseQA\\
HourVideo & \faSunO  & (20min, 120min) & 12976 & - & \ding{51} & CloseQA\\
EgoLifeQA& mostly \faSunO & 44.3 h & 6000 & 5 & \ding{51} & CloseQA\\
\midrule
\textbf{EgoNight-VQA} & Aligned \faSunO \& \faMoonO & (24s, 214s) & 3658 & 12 & \ding{51} & OpenQA \\
\bottomrule
\end{tabular}}
\caption{Comparison between EgoNight-VQA and prior egocentric VQA benchmarks. \faSunO\ means dayytime, while \faMoonO \  indicates nighttime. }
\label{tab:benchmark_comparison}
\end{table*}

\subsection{More EgoNight-VQA Explanations and Examples}
\subsubsection{QA Type Defination}
\label{sec:qa_type_def}
We present the 12 QA types with their detailed definitions in Tab.~\ref{tab:vqa_types_detail}.

\begin{table}[h]
\centering
\caption{\textbf{Detailed descriptions of the 12 QA types in EgoNight-VQA.} 
Paired QA types share the same QAs across day–night counterparts, 
while unpaired QA types are evaluated only on nighttime videos.}
\label{tab:vqa_types_detail}
\resizebox{\textwidth}{!}{
\begin{tabular}{lll}
\toprule
\textbf{QA Type} & \textbf{Attribute} & \textbf{Description} \\
\midrule
Object Recognition & Paired & Identify and recognize specific objects in the scene (e.g., ``What is on the table?''). \\
Text Recognition & Paired & Read and interpret visible text or logos (e.g., ``What does the sign say?''). \\
Spatial Reasoning & Paired & Understand spatial relations between objects (e.g., ``What is left of the chair?''). \\
Scene Sequence & Paired & Recall the temporal order of visited scenes (e.g., ``Which room did I enter after the kitchen?''). \\
Navigation & Paired & Working as an navigation assistant after watched the whole video (e.g., ``How can I reach place B from place A?''). \\
Counting of Statics & Paired & Count static objects visible in the scene (e.g., ``How many chairs are in the room?''). \\
Action Recognition & Paired & Identify human actions or interactions (e.g., ``What action is being performed?''). \\
Non-Common-Sense Reasoning & Paired & Judge unusual or physically implausible cases, for synthetic videos. (e.g., ``Is the door embedded inside the wall?''). \\
\midrule
Lighting Recognition & Unpaired & Recognize the illumination source, also include counting. (e.g., ``How many light sources are in the room?''). \\
Lighting Change & Unpaired & Detect changes in lighting conditions (e.g., ``Did the light turn off during the clip?''). \\
Dynamic Detection & Unpaired & Detect dynamic moving objects (e.g., ``Is a car/person moving across the scene?''). \\
Counting of Dynamics & Unpaired & Count the number of dynamic objects or events (e.g., ``How many people walked by?''). \\
\bottomrule
\end{tabular}}
\end{table}

\subsubsection{QA Examples}\label{supp:qa_examples}
\begin{figure}[t!]
    \centering
    \includegraphics[width=\linewidth]{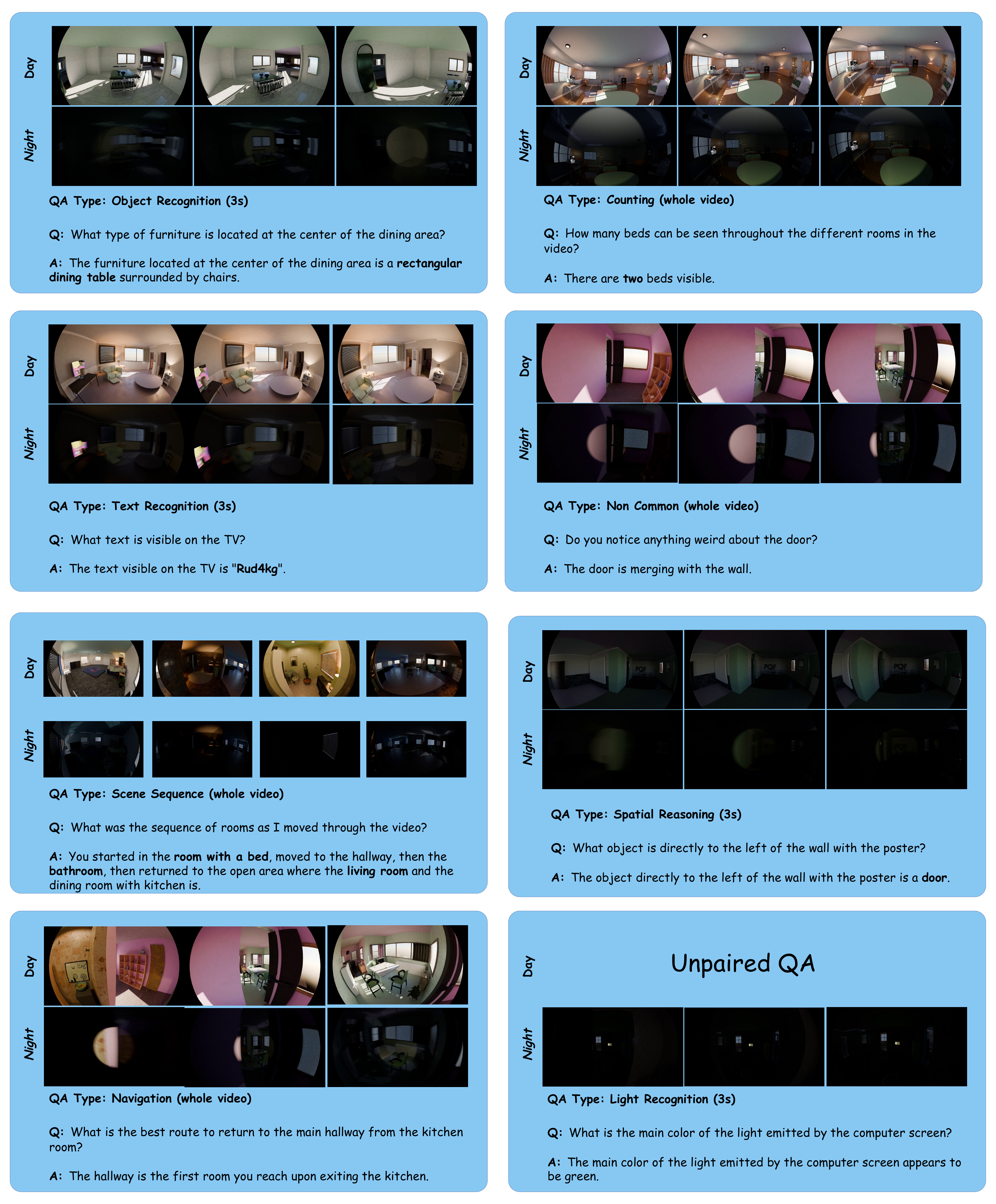}
    \caption{
     More QA examples from EgoNight-Synthetic dataset. 
    }
    \label{fig:qa_ex_syn}
    \vspace{-0.15in}
\end{figure}

\begin{figure}[t!]
    \centering
    \includegraphics[width=0.85\linewidth]{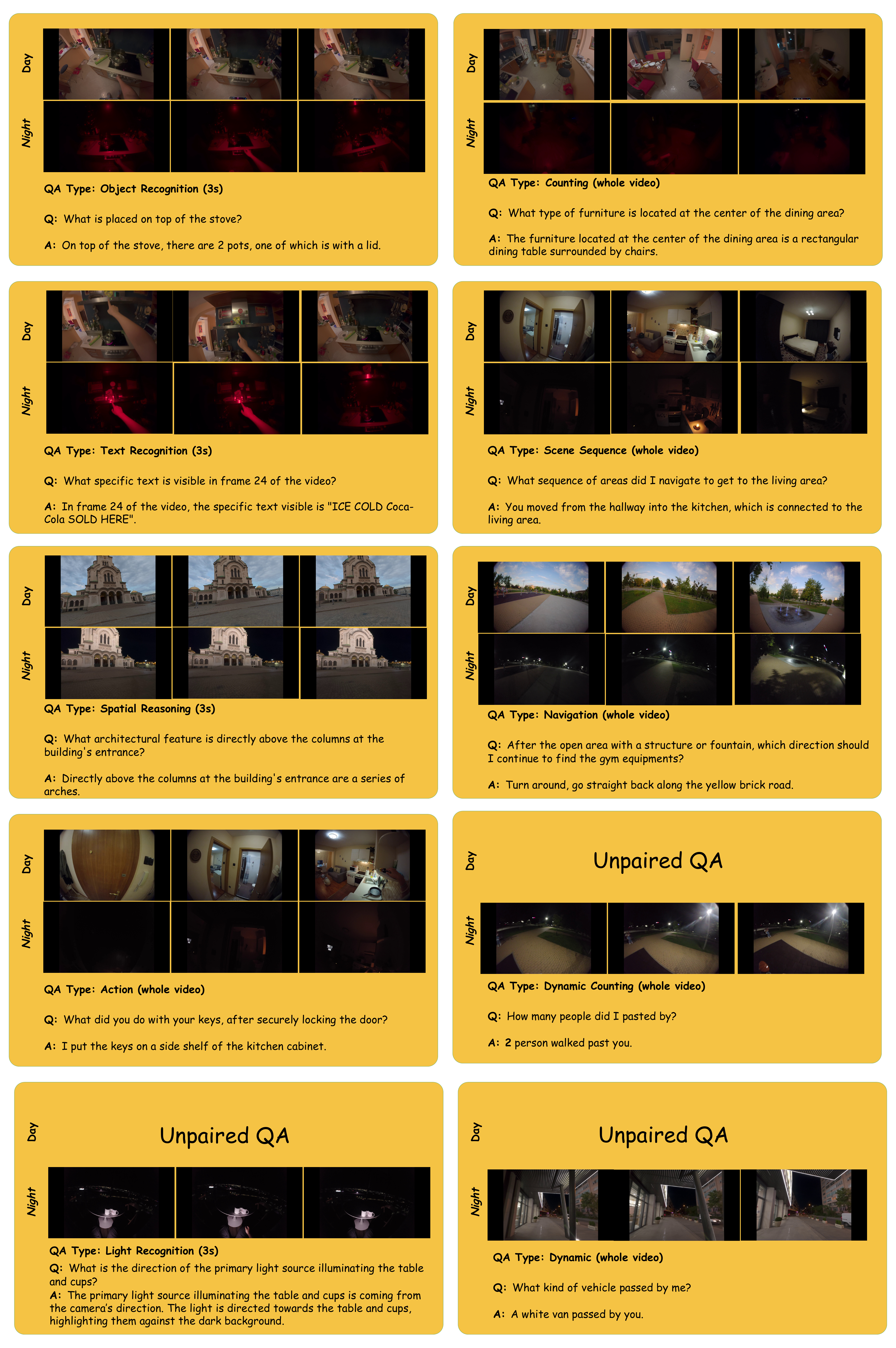}
    \caption{
     More QA examples from EgoNight-Sofia dataset. 
    }
    \label{fig:qa_ex_sofia}
\end{figure}

\begin{figure}[t!]
    \centering
    \includegraphics[width=0.90\linewidth]{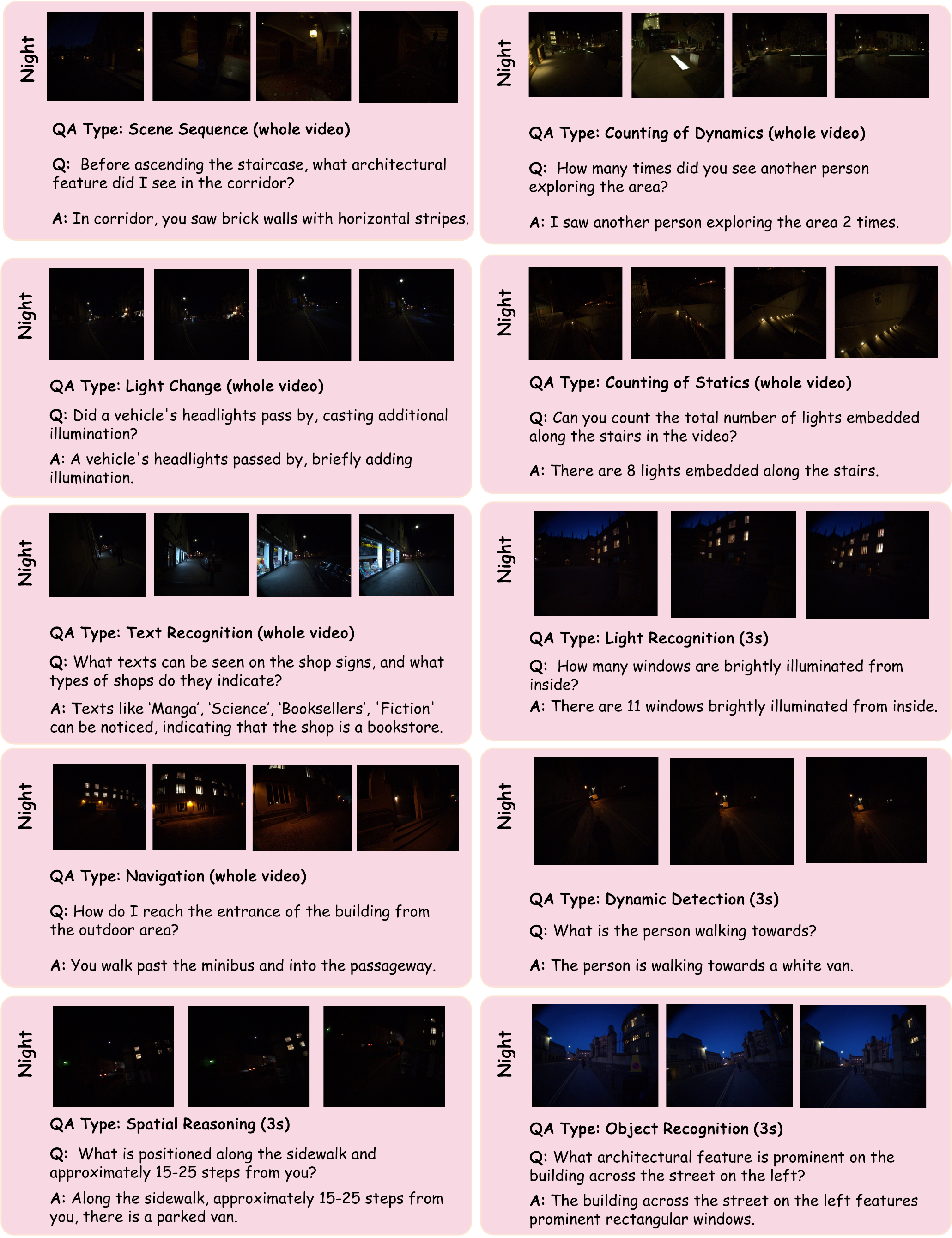}
    \caption{
     More QA examples from EgoNight-Oxford dataset. 
    }
    \label{fig:qa_ex_oxford}
\end{figure}

We show more QA examples of EgoNight-Synthetic in Fig.~\ref{fig:qa_ex_syn}, EgoNight-Sofia in Fig.~\ref{fig:qa_ex_sofia}, and EgoNight-Oxford in Fig.~\ref{fig:qa_ex_oxford}. For those paired QA types, we show both day and night frames, while for those unpaired QA types, we demonstrate nighttime frames only.  Three frames are shown if the QA is spatial or static related, while more frames are given if the QA is temporal or more dynamic related. 

\subsection{More Experiments Setups}
\label{sec:more_experiment_setup}
\begin{table}[h!]
\centering
\begin{tabular}{c|c}
\hline
Model & Inference Speed (min) \\ \hline
GPT-4.1 & $<$5 \\ \hline
Gemini 2.5 Pro & $<$5 \\ \hline
InternVL3-8B & $<$5 \\ \hline
Qwen2.5-VL-72B & 25 \\ \hline
Qwen2.5-VL-7B & $<$5 \\ \hline
Qwen2.5-VL-3B & $<$5 \\ \hline
GLM-4.1V-9B-Base & $<$5 \\ \hline
VideoLLaMA3-7B & $<$5 \\ \hline
LLaVA-NeXT-Video-7B & 50 \\ \hline
EgoGPT & $<$5 \\ \hline
\end{tabular}
\caption{Inference speed for different models (per Video).}
\label{tab:inference_speed}
\end{table}
\subsubsection{Setups for EgoNight-VQA Experiments.}
In this section, we describe the model setup, how GPT is used as the judge, the prompting strategy, the GPU resources, and the approximate runtime for each dataset. 
For the closed-source model, we directly use the API call. For open source models, we use LLama-Factory~\cite{zheng2024llamafactory} except VideoLLama3~\cite{zhang2023video} and EgoGPT~\cite{yang2025egolife}. We use NVIDIA A6000 GPUs for all the model, except for Qwen2.5-VL-72B, we use 2 NVIDIA H200 GPUs for larger GPU memory. The inference speed for each model is shown in Tab.~\ref{tab:inference_speed}. Video frames are sampled at 2 fps for EgoNight-Synthetic, and 1 fps for EgoNight-Sofia and EgoNight-Oxford, without imposing a maximum frame limit. To further ensure fairness and consistency, the exact prompts used for each task are provided below. 

\noindent\textbf{Model Evaluation Prompt.} For evaluating the language model, we use the following prompt:

\begin{minipage}{\textwidth}
\begin{lstlisting}[basicstyle=\small\itshape]
Please carefully read the question, use the visual cues in the {video} to answer the question: {question}. 

The original FPS of the video is {original_video_fps}. This image set is obtained by sampling at {sampling} fps. 

Do not include any other content. You need to answer the question in any case and not demand additional context information.
Note: All the actions mentioned refer to the person who recorded the video.
\end{lstlisting}
\end{minipage}

\noindent\textbf{Evaluation Protocol.} Since the questions and answers are open-ended, we utilize GPT-4.1~\cite{achiam2023gpt} as a judge. Here is the prompt for evaluating the score given the model prediction, the ground truth answer, and the corresponding question:

\begin{minipage}{\textwidth}
\begin{lstlisting}[basicstyle=\small\itshape]
role: system,
content: You are an intelligent chatbot designed for evaluating the correctness of AI assistant predictions for question-answer pairs.
Your task is to compare the predicted answer with the ground-truth answer and determine if the predicted answer is correct or not. Here's how you can accomplish the task:
INSTRUCTIONS:
1. Focus on the correctness and accuracy of the predicted answer with the ground-truth.
2. Consider uncertain predictions, such as 'it is impossible to answer the question from the video', as incorrect, unless the ground truth answer also says that.
role: user,
content: Please evaluate the following video-based question-answer pair:
Question: {question}
Ground truth correct Answer: {answer}
Predicted Answer: {predicted_answer}
Provide your evaluation as a correct/incorrect prediction along with the score where the score is an integer value between 0 (fully wrong) and 5 (fully correct). The middle score provides the percentage of correctness. For question that counting the number of objects, if the predicted answer fells in the range of the ground truth answer, it should be considered as correct.
Please generate the response in the form of a Python dictionary string with keys 'pred', 'score' and 'reason', where value of 'pred' is a string of 'correct' or 'incorrect',
value of 'score' is in INTEGER, not STRING and value of 'reason' should provide the reason behind the decision."
\end{lstlisting}
\end{minipage}
To further validate the LLM-as-a-Judge strategy, we divide all annotations into two groups: (a) answers verified but not modified by humans (preserving the GPT style) and (b) answers modified or created by humans. The ratio is approximately $4:6$, indicating a high human modification rate.  When evaluating the accuracy of GPT-4.1 separately in the two subsets, we obtain $26.73\%$ for group (a) and $27.87\%$ for group (b). The similar scores show that GPT-as-Judge does not prefer GPT-generated answers over human-authored ones. Further more, We randomly sampled 300 QA pairs (questions, ground truth, model answers, and the corresponding LLM-assigned scores) and asked human evaluators to judge whether each score from the LLM was correct. This yielded an agreement rate of $95.67\%$, indicating strong alignment between human judgment and the LLM-as-a-Judge decisions, and thus demonstrating the reliability of the LLM-based evaluation.

\subsubsection{Setups for Day-Night Correspondence Retrieval.}

In this section, we describe the setup of the model, method, e.g. how feature-based retrieval for vision encoders, how prompt VLMs, metric, result, GPU, cost time, and other details.\\
\textbf{i) Spatial Retrieval (Place Recognition).}  
For feature-based methods~\cite{oquab2023dinov2, bolya2025perceptionencoderbestvisual}, we calculate the CLS tokens $f^i$ of each frame within the video clip with the vision encoder, with $i$ the frame index in the clip. 
Then, the "best matching" strategy is implemented to calculate the similarity between the query clip $v_q$ and the database clip $v_d$.
The best cosine similarity between the features of the query clip $f^i_q$ and the database clip $f^j_d$,
\begin{equation}
    \sigma(v_q, v_d) = \max_{i\in[0, s-1], \space j\in[0, s-1]}cos(f^i_q, f^j_d).
\end{equation}
The database video clips are ordered based on the similarity $\sigma$ and then the most similar clip is retrieved. 

Similarly, for MLLM-based methods~\cite{achiam2023gpt, chen2024internvl}, we ask the MLLM to assess the "pairwise" similarity between each query-database pair and order the database clips by similarity. 
The prompt to the MLLM is as follows:

\begin{minipage}{\textwidth}
\begin{lstlisting}[basicstyle=\small\itshape]
You are given two video clips from different scenes. 
Your task is to evaluate how similar these two scenes are based on their spatial layout, furniture, objects, architectural features, and overall room structure.

CLIP STRUCTURE:
- Images 1->(s-1) from Query Scene 
- Images s->{2s-1} from Database Scene

TASK:
Please carefully analyze and compare the spatial layout, furniture placement, objects, architectural features, and overall room structure between these two video clips.

IMPORTANT: Please respond with ONLY a single numerical similarity score between 0.0 and 1.0, where:
- 0.0 = Completely different scenes (different rooms/locations)
- 1.0 = Identical or nearly identical scenes (same room/location)
- Values in between represent varying degrees of similarity

Example responses: "0.85", "0.23", "0.67"
1.0 should be used when the two scenes are identical, so don't use 1.0 if the two scenes are not 100% identical.
Please provide only the numerical score without any additional text or explanation.
\end{lstlisting}
\end{minipage}

It is noticeable that existing MLLMs have difficulty in processing long-horizon and multi-scene videos. 
We also conduct the ``all-in-one-prompt" experiments by inputting all the images of the query clip and the database clips in one prompt and asking the MLLM to output the ordered database clips. 
The ``all-in-one-prompt" strategy leads to largely degraded performance, as shown in Tab.~\ref{tab:abation_prompt}.
\begin{table}[h]
\centering
\begin{tabular}{c|c|c}
\hline
& \multicolumn{2}{c}{Spatial Retrieval R@1 - Synthetic} \\ 
Prompt Strategy & Day $\rightarrow$ Day & Night $\rightarrow$ Day \\ \hline
Pairwise        &        75.6               &          54.1              \\ \hline
All-in-one      &         10.5              &             28.5           \\ \hline
\end{tabular}
\caption{Ablation on prompting for night-to-day spatial retrieval task.}\label{tab:abation_prompt}
\end{table}

\textbf{ii) Temporal Localization.}  
The mIoU metric is defined as:  
\begin{equation}
\text{mIoU} = \frac{1}{M} \sum_{m=1}^M 
\frac{\big|[t_i, t_j] \cap [t_i^*, t_j^*]\big|}
     {\big|[t_i, t_j] \cup [t_i^*, t_j^*]\big|},
\end{equation}
where $M$ denotes the total number of meta-tasks (1000 in our setup). 
For feature-based temporal localization, we apply the "best-match" strategy similar as spatial localization, localizing the query clip to the frame stamp with the best clip-to-clip similarity:
\begin{equation}
    i = \arg\max_{i}\sigma(v_q, v_d),\space v_d = v_D[i:i+s], 
\end{equation}

$v_D$ is the parent full video and the end frame will be $i+s-1$. 
For MLLM-based method, we input the query clip and the parent full video in the prompt and ask the MLLM to output the start and end frame of the query within the full video. The prompt is as follows:

\begin{minipage}{\textwidth}
\begin{lstlisting}[basicstyle=\small\itshape]
You are given a query video clip and a complete video sequence from the same scene. 
Your task is to find the exact temporal position where the query clip appears in the complete video sequence.
IMPORTANT CONTEXT:
- The query clip shows s consecutive frames from a video sequence
- The complete video sequence shows ALL frames from the same scene in chronological order
- The query clip appears as a consecutive subsequence somewhere within the complete video sequence
- You need to find the exact start and end frame numbers where this subsequence appears

IMAGE STRUCTURE:
- Images 1->s: Query video clip (consecutive frames to find)
- Images {s+1}-{s+1+video_len}: Complete video sequence (all frames in chronological order)

TOTAL IMAGES: {query_count + database_count} images

TASK:
1. Look at the query clip to understand what sequence you're looking for
2. Search through the complete video sequence to find where this exact sequence appears
3. The query sequence should appear as consecutive frames in the complete video sequence
4. Pay attention to camera movements, object positions, and scene changes to identify the matching sequence

FRAME NUMBERING:
- The complete video sequence frames are numbered from {min(database_frame_numbers)} to {max(database_frame_numbers)}
- You need to return the actual frame numbers from this range

RESPONSE FORMAT:
Respond with ONLY two numbers separated by a comma: "start_frame,end_frame"
- start_frame: The frame number where the query clip begins in the complete video sequence
- end_frame: The frame number where the query clip ends in the complete video sequence

Example: If the query clip appears at frames 15-19 in the complete sequence, respond: "15,19"
Valid frame range: {min(database_frame_numbers)} to {max(database_frame_numbers)}
\end{lstlisting}
\end{minipage}

\subsubsection{Setups for Egocentric Depth Estimation at Night.}
We evaluate four off-the-shelf monocular depth systems without night-specific fine-tuning. For each, we highlight features pertinent to our setting. \textcolor{green!60!black}{(\textbf{F})} denotes support for fisheye egocentric images; \textcolor{blue}{(\textbf{U})} denotes undistorted/pinhole images.
\begin{enumerate}
  \item \textbf{Depth Anything V2 (metric). \textcolor{blue}{(\textbf{U})}} Foundation MDE model (DPT head with DINOv2 backbone) trained on large-scale synthetic labels plus pseudo-labeled real images. We use the \emph{official metric} checkpoints: \textit{Indoor} (Hypersim-tuned) for indoor frames and \textit{Outdoor} (VKITTI2-tuned) for outdoor frames. Outputs metric depth in meters and is known for strong zero-shot generalization. 
  \item \textbf{StreamVGGT. \textcolor{blue}{(\textbf{U})}} A causal/streaming transformer for video geometry that processes frames sequentially with state caching to improve temporal consistency and enable real-time inference. We run it in streaming mode to obtain per-frame depth on egocentric sequences.
  \item \textbf{Depth Any Camera (DAC). \textcolor{green!60!black}{(\textbf{F})}} Zero-shot \emph{metric} depth across diverse camera models via a unified ERP (equirectangular) representation with pitch-aware image-to-ERP conversion and FoV alignment. We use the official release with default settings on our pinhole inputs.
  \item \textbf{UniK3D. \textcolor{green!60!black}{(\textbf{F})}} Universal-camera monocular 3D estimation with a spherical 3D formulation and a learned “pencil-of-rays” camera module, enabling accurate metric depth across pinhole, fisheye, and panoramic views. We run the official model in eval mode; when available, we provide intrinsics for pinhole frames.
\end{enumerate}

\subsection{More Experimental Results}
\label{sec:more_exp_result}
\begin{table*}[t]
\centering
\small
\renewcommand{\arraystretch}{1.15}
\resizebox{\textwidth}{!}{
\begin{tabular}{lcccccccccc}
\toprule
\textbf{Model} &
\textbf{Object Rec.} &
\textbf{Text Rec.} &
\textbf{Spatial} &
\textbf{Scene Seq.} &
\textbf{Nav.} &
\textbf{Light Rec.} &
\textbf{Counting} &
\textbf{Non-Common} &
\textbf{Overall} \\
\midrule
\multicolumn{10}{l}{\emph{Closed-Source MLLMs}} \\
\midrule
Gemini       & 25.94 & 39.39 & 32.43 & 35.47 & 30.77 & 31.97 & 21.88 & 15.15 & 28.34 \\
GPT-4.1      & 25.19 & 54.55 & 35.42 & 28.44 & 27.09 & 35.25 & 20.83 & 16.67 & 27.75 \\
\midrule
\multicolumn{10}{l}{\emph{Open-Source MLLMs}} \\
\midrule
 InternVL3-8B    & 17.29 & 10.61 & 28.34 & 20.80 & 10.37 & 18.03 & 16.93 & 21.21 & 18.97 \\
Qwen2.5-VL-72B  & 15.41 & 16.67 & 28.88 & 21.41 &  7.02 & 12.30 & 10.94 & 21.21 & 17.15 \\
Qwen2.5-VL-7B      &  6.77 & 13.64 & 17.98 & 11.93 &  9.03 & 15.57 & 14.06 & 18.69 & 13.26 \\
Qwen2.5-VL-3B      &  7.89 & 22.73 & 17.17 & 14.37 & 11.37 &  8.20 & 13.02 & 12.63 & 13.06 \\
GLM-4.1V-9B-Base    & 13.16 & 36.36 & 23.71 & 21.71 &  7.02 & 15.57 & 19.27 & 14.14 & 17.69 \\
LLaVA-NeXT-Video-7B  &  5.26 & 10.61 & 10.08 &  4.59 &  3.01 & 16.39 &  4.69 &  9.60 &  6.85 \\
VideoLLaMA3-7B & 10.90 & 21.21 & 19.07 & 23.55 &  7.02 &  7.38 & 18.49 & 16.67 & 15.97 \\
\midrule
\multicolumn{10}{l}{\emph{Egocentric MLLMs}} \\
\midrule
EgoGPT       &  6.02 & 19.70 & 18.53 & 19.88 &  8.36 &  8.20 & 17.71 & 17.68 & 14.79 \\
\midrule
\midrule
\multicolumn{10}{l}{\emph{Average across all models}} \\
\midrule
Average      & 13.38 & 24.55 & 23.16 & 20.21 & 12.11 & 16.89 & 15.78 & 16.36 & 17.38 \\
\bottomrule
\end{tabular}
}
\caption{Night-time VQA accuracy (\%) per model across all QA categories for EgoNight-Synthetic.}
\vspace{-0.15in}
\label{tab:night-acc-by-task-synthetic}
\end{table*}

\begin{table*}[h]
\centering
\small
\renewcommand{\arraystretch}{1.15}
\resizebox{\textwidth}{!}{
\begin{tabular}{lcccccccccccc}
\toprule
\textbf{Model} &
\textbf{Object Rec.} &
\textbf{Text Rec.} &
\textbf{Spatial} &
\textbf{Scene Seq.} &
\textbf{Action} &
\textbf{Nav.} &
\textbf{Light Rec.} &
\textbf{Counting} &
\textbf{Dyn. Light} &
\textbf{Dynamic} &
\textbf{Dyn. Count} &
\textbf{Avg.} \\
\midrule
\multicolumn{13}{l}{\emph{Closed-Source MLLMs}} \\
\midrule
GPT-4.1      & 24.44 & 33.78 & 41.32 & 30.09 & 38.10 & 27.27 & 38.33 & 24.62 & 30.00 & 13.33 & 25.00 & 31.06 \\
Gemini       & 32.22 & 47.30 & 35.54 & 24.78 & 34.92 & 29.29 & 43.33 & 27.69 & 15.00 & 26.67 & 40.00 & 32.67 \\
\midrule
\multicolumn{13}{l}{\emph{Open-Source MLLMs}} \\
\midrule
InternVL3-8B      & 16.67 & 17.57 & 33.88 & 24.78 & 22.22 & 21.21 & 20.00 & 22.31 & 25.00 &  6.67 & 15.00 & 22.61 \\
Qwen2.5-VL-72B  & 14.44 & 21.62 & 36.36 & 18.58 & 19.05 & 25.25 & 18.33 & 13.85 & 20.00 &  6.67 & 20.00 & 20.99 \\
Qwen2.5-VL-7B      &  7.78 & 10.81 & 21.49 & 18.58 & 11.11 & 18.18 &  1.52 & 15.38 &  9.52 &  6.25 & 15.00 & 14.02 \\
Qwen2.5-VL-3B      & 11.11 &  8.11 & 23.97 & 13.27 & 11.11 & 16.16 & 10.00 & 15.38 &  5.00 & 13.33 & 10.00 & 14.16 \\
GLM-4.1V-9B-Base        & 12.22 & 12.16 & 30.58 & 15.04 & 14.29 & 17.17 & 11.67 & 25.38 & 15.00 &  6.67 & 10.00 & 18.14 \\
VideoLLaMA3-7B &  3.33 &  5.41 & 13.22 & 11.50 &  6.35 &  9.09 &  8.33 & 18.46 & 10.00 & 20.00 & 15.00 & 10.68 \\
LLaVA-NeXT-Video-7B    &  8.89 &  5.41 & 18.18 & 12.39 & 12.70 & 15.15 & 11.67 & 13.85 &  0.00 & 13.33 & 20.00 & 12.67 \\

\midrule
\multicolumn{13}{l}{\emph{Egocentric MLLMs}} \\
\midrule
EgoGPT       &  9.18 &  3.41 & 19.26 & 16.54 &  6.67 &  7.96 & 10.00 & 14.97 &  0.00 &  0.00 & 10.00 & 11.47 \\
\midrule
\multicolumn{13}{l}{\emph{Average across all models}} \\
\midrule
Average      & 13.99 & 16.31 & 27.29 & 18.53 & 17.45 & 18.53 & 17.16 & 19.13 & 12.94 & 11.26 & 18.00 & 18.76 \\
\bottomrule
\end{tabular}
}
\caption{Night-time VQA accuracy (\%) per model across all QA categories for EgoNight-Sofia.}
\vspace{-0.15in}
\label{tab:night-acc-by-task-sofia}
\end{table*}

\begin{table*}[h!]
\centering
\small
\renewcommand{\arraystretch}{1.15}
\resizebox{1.\textwidth}{!}{
\begin{tabular}{lcccccccccccc}
\toprule
\textbf{Models} &
\textbf{Object Rec.} &
\textbf{Text Rec.} &
\textbf{Spatial} &
\textbf{Scene Seq.} &
\textbf{Action} &
\textbf{Nav.} &
\textbf{Light Rec.} &
\textbf{Counting} &
\textbf{ Dyn. Light} &
\textbf{Dynamic} &
\textbf{Dyn. Count} &
\textbf{Avg.} \\
\midrule
\multicolumn{13}{l}{\emph{Closed-Source MLLMs}} \\
\midrule
GPT-4.1 & 64.52 & 34.88 & 41.35 & 34.13 & 65.59 & 43.75 & 30.43 & 18.49 & 18.52 & 37.84 & 18.52 & 38.95 \\
Gemini 2.5 Pro    & 56.14 & 46.51 & 38.30 & 27.27 & 59.55 & 32.65 & 31.11 & 18.97 & 23.33 & 37.14 &  3.70 & 34.83 \\
\midrule
\multicolumn{13}{l}{\emph{Open-source MLLMs}} \\
\midrule
InternVL3-8B      & 40.00 & 27.91 & 18.68 & 14.66 & 36.78 & 20.83 &  6.98 &  9.91 &  6.67 & 37.14 &  7.41 & 20.57 \\
Qwen2.5-VL-72B    & 38.18 & 39.53 & 29.67 & 13.79 & 22.99 & 23.96 & 16.28 & 17.12 & 16.67 & 11.43 & 11.11 & 22.07 \\
Qwen2.5-VL-7B     &  9.09 & 23.26 & 20.88 & 11.21 & 10.34 & 15.62 &  9.30 & 11.71 &  3.33 & 11.43 & 18.52 & 13.35 \\
Qwen2.5-VL-3B     & 14.55 & 13.95 &  9.89 & 13.79 & 19.54 & 20.83 &  6.67 & 10.81 &  3.70 & 17.14 &  6.98 & 13.62 \\
GLM-4V            & 22.81 & 30.23 & 21.43 & 15.52 & 28.74 &  9.38 & 19.57 & 17.12 &  6.67 & 37.14 & 14.81 & 19.57 \\
VideoLLaMA3-7B    & 20.00 & 11.63 & 15.38 &  9.57 & 10.34 &  7.29 &  9.52 &  9.01 &  0.00 & 17.65 &  7.41 & 10.81 \\
LLaVA-NeXT-Video-7B & 9.09 &  4.65 & 10.99 &  0.00 &  0.00 &  0.00 &  6.98 &  0.00 &  0.00 &  2.86 &  0.00 &  2.86 \\

\midrule
\multicolumn{13}{l}{\emph{Egocentric MLLMs}} \\
\midrule
EgoGPT            & 21.13 & 27.08 & 14.14 &  3.42 & 14.61 &  6.25 & 11.11 &  6.25 & 21.62 & 18.92 & 17.86 & 12.44 \\
\midrule
\multicolumn{13}{l}{\emph{Average across all models}} \\
\midrule
Avg.              & 29.81 & 25.98 & 22.32 & 14.55 & 27.16 & 18.09 & 14.96 & 12.01 & 10.75 & 22.95 & 10.33 & 19.04 \\
\bottomrule
\end{tabular}
}
\caption{Night-time VQA accuracy (\%) per model across all QA categories for EgoNight-Oxford.}
\vspace{-0.15in}
\label{tab:night-acc-by-task-oxford}
\end{table*}

In this section, we show models per QA accuracy on each dataset  for EgoNight-Synthetic in Tab.~\ref{tab:night-acc-by-task-synthetic}, EgoNight-Sofia in Tab.~\ref{tab:night-acc-by-task-sofia}, and EgoNight-Oxford in Tab.~\ref{tab:night-acc-by-task-oxford}. 

\subsection{More Visualization Results}
\subsubsection{EgoNight-VQA}
\label{supp:more_visualization}
\begin{figure}[h]
    \centering
    \includegraphics[width=\linewidth]{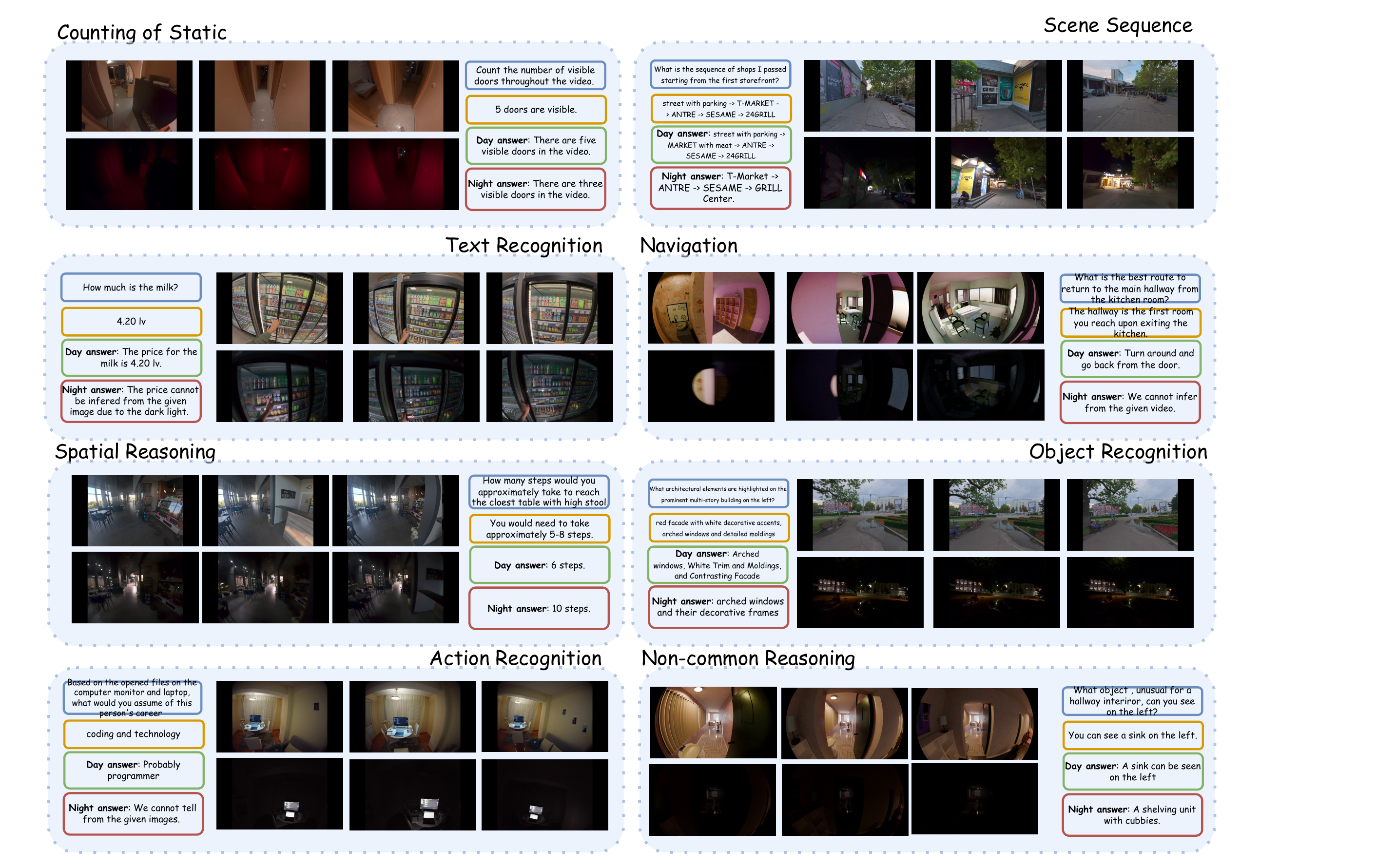}
    \caption{
     More QA examples with day and night answer produced by the same model. 
    }
    \label{fig:qa_gap}
    \vspace{-0.15in}
\end{figure}

\begin{figure}[h]
    \centering
    \includegraphics[width=\linewidth]{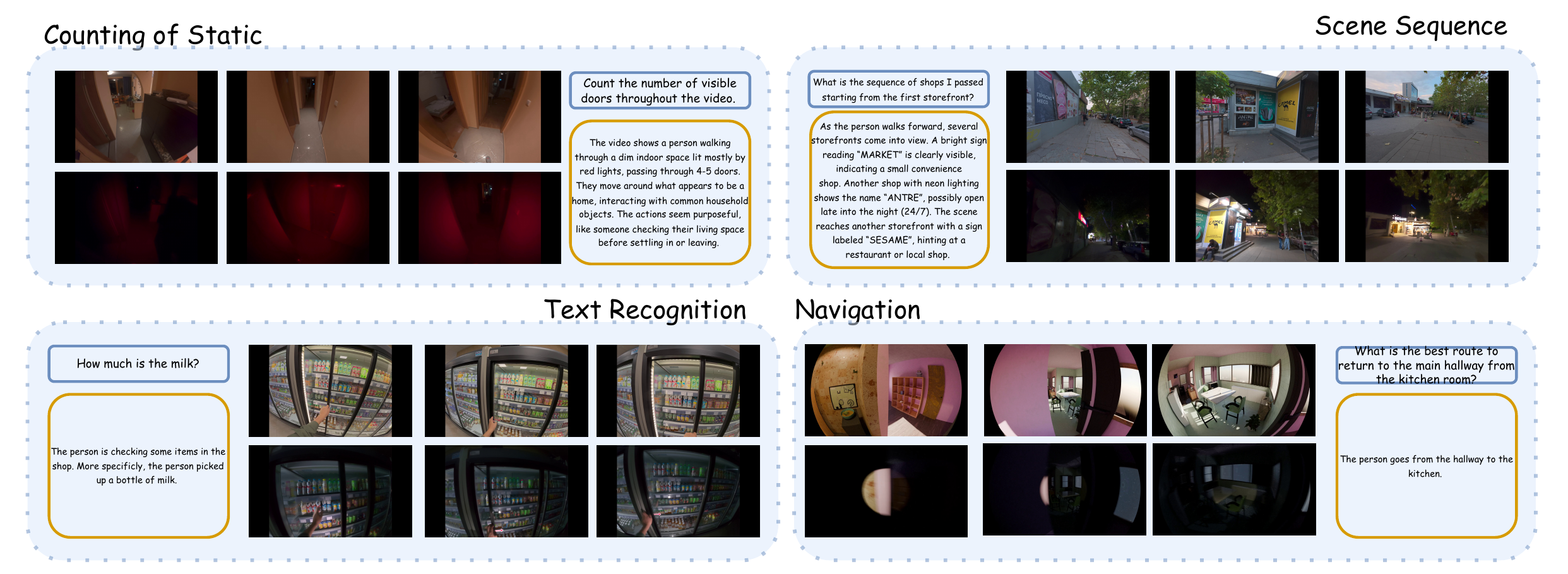}
    \caption{
     More examples shows the caption together with generated questions. 
    }
    \label{fig:caption}
    \vspace{-0.15in}
\end{figure}

We further investigate failure reasons by systematically altering illumination, motion blur, and camera noise (sampling quality) in synthetic videos, while keeping scene, annotations, and trajectory fixed in the synthetic dataset.  The procedure of generating such difficulty level is described in Appendix.~\ref{sec:more_construction}, we calculate the averaged accuracy from Tab.~\ref{tab:main_result} as $18.47\%$ for Easy (moderately dark), $15.88\%$ for Medium (very dark with camera noise), $12.95\%$ for Hard (as dark as medium, with motion blur).  From the results, we highlight that: ii) Lower illumination together with camera sensor noise leads to the most significant drop due to loss of contrast and missing fine details. ii) Motion blur further harms performance by causing temporal ambiguity and object shape distortion. iii) These results confirm that night conditions VQA is challenging, highlighting the need for specialized night-egocentric benchmarks. We also visualize more failure cases, and give  more analysis in Appendix.~\ref{supp:more_visualization}.
Here we provide more failure cases, which compare the day and night VQA output in Fig.~\ref{fig:qa_gap}. This clearly shows the gap between the day and night video understanding. Also, we provide examples of question generated and the corresponding caption to show caption reasoning ability in Fig.~\ref{fig:caption}. Here we provide a more detailed analysis of the failure reason:
\begin{itemize}
    \item \textbf{Extreme Illumination (Counting – Static):} Due to strong red lighting, two doors in the scene become nearly invisible, causing the model to undercount objects.

    \item \textbf{Small Object Disappearance (Text Recognition):} The price tag becomes too small and poorly illuminated at night, making it unreadable and leading to text recognition failure.

    \item \textbf{Spatial Confusion from Limited View (Navigation):} Restricted field of view at night hides key spatial cues (e.g., landmarks, corridor orientation), causing incorrect navigation decisions.

    \item \textbf{Motion Blur (Action Recognition):} Fast hand movement introduces motion blur, making the model misinterpret the content displayed on the screen.
\end{itemize}

\subsubsection{Day-Night Correspondence Retrieval}
\begin{figure}[h]
\centering
\includegraphics[width=0.9\textwidth]{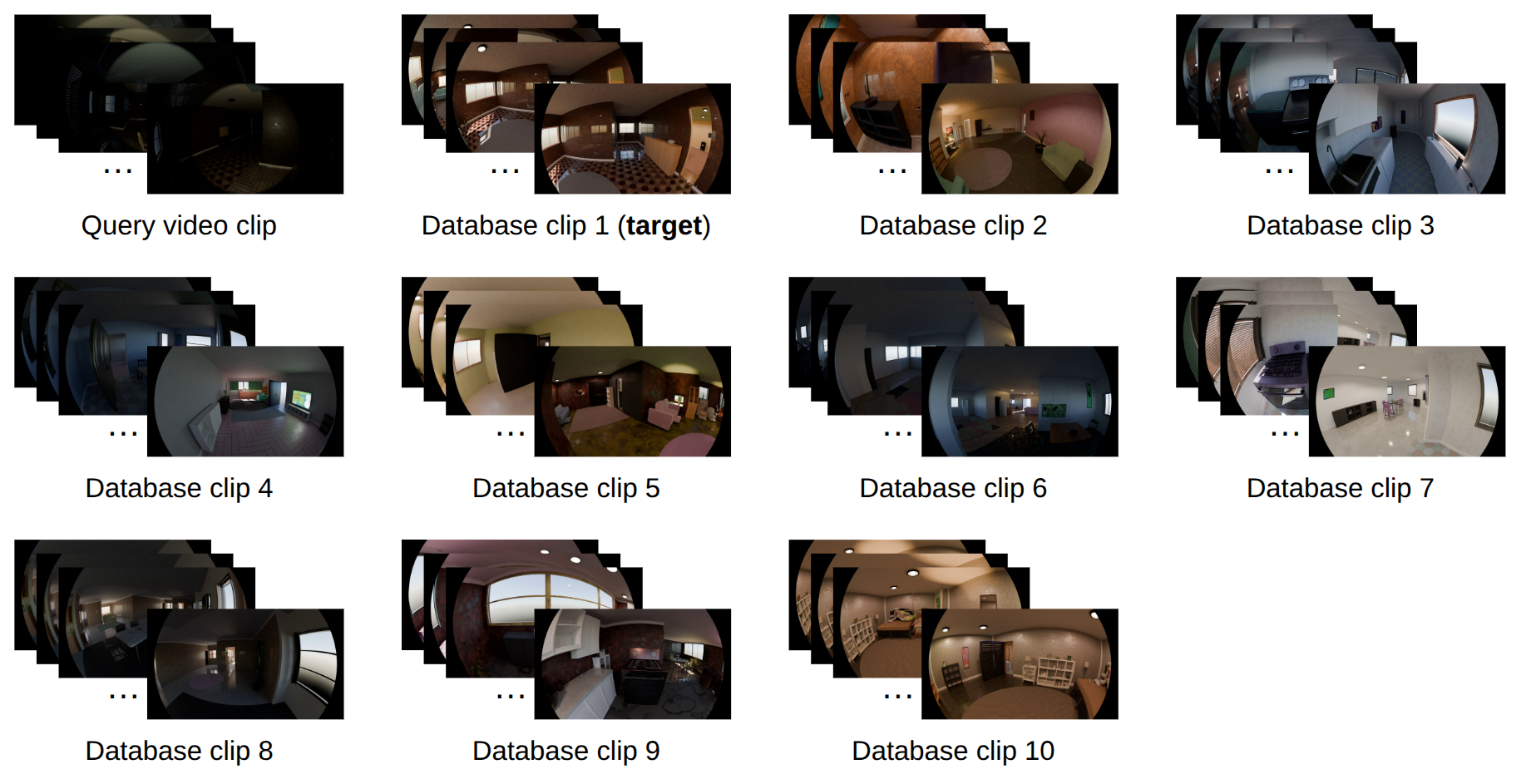} 
\begin{tabular}{c|c|c|c|c|c|c|c|c|c|c}
\hline
Methods & DB clip 1 & DB 2 & DB 3 & DB 4 & DB 5 & DB 6 & DB 7 & DB 8 & DB 9 & DB 10 \\ \hline
DINOv2       & 0.67 &  0.66 & 0.58 & 0.54 & 0.67 & \color{red}{\textbf{0.70}} & 0.61 & 0.66 & 0.59 &  0.67 \\ \hline
Percep. Enc. & 0.83 &  0.82 & 0.78 & 0.81 &  0.81  & 0.80  &  0.79  & 0.81  & 0.78  &  \color{red}{\textbf{ 0.85}}  \\ \hline
GPT-4.1      &  \color{green}{\textbf{0.92}}  & 0.72 & 0.62 & 0.18 & 0.18 & 0.18 & 0.18 & 0.12 & 0.12 & 0.12 \\ \hline
InternVL 8B  & \color{green}{\textbf{0.55}} & 0.32 & 0.32  & 0.30 & 0.30 & 0.25 & 0.25 & 0.21 & 0.15 & 0.15 \\ \hline
\end{tabular}
\caption{Qualitative Result on one meta sample of spatial retrieval. The query video clip and the database video clips are visualized in the image. The table below the figure shows the similarity score between the query and the database clips calculated with different methods. The most similar one is in \textbf{bold}, and correct retrieval is in {\color{green}\textbf{green}}, and the incorrect one is in  {\color{red}\textbf{red}}.} \label{fig: qualitative_spatial_retrieval}
\vspace{-0.15in}
\end{figure}

We visualize the qualitative result on one meta sample of Night-to-Day spatial retrieval to better demonstrate the experiment setup and the performance of the benchmarked methods. 
As shown in Fig.~\ref{fig: qualitative_spatial_retrieval}, the light condition of the query video clip is drastically different from that of the database clips. 
Such a difference imposes a great challenge for existing methods in distinguishing the target scenes from the other candidate databases' clips, showing the value of the dataset in the place recognition task.

\subsubsection{Egocentric Depth Estimation at Night}
\label{sec:EgoDepthQualitative}
\begin{figure}[h!]
    \centering
    \includegraphics[width=0.9\linewidth]{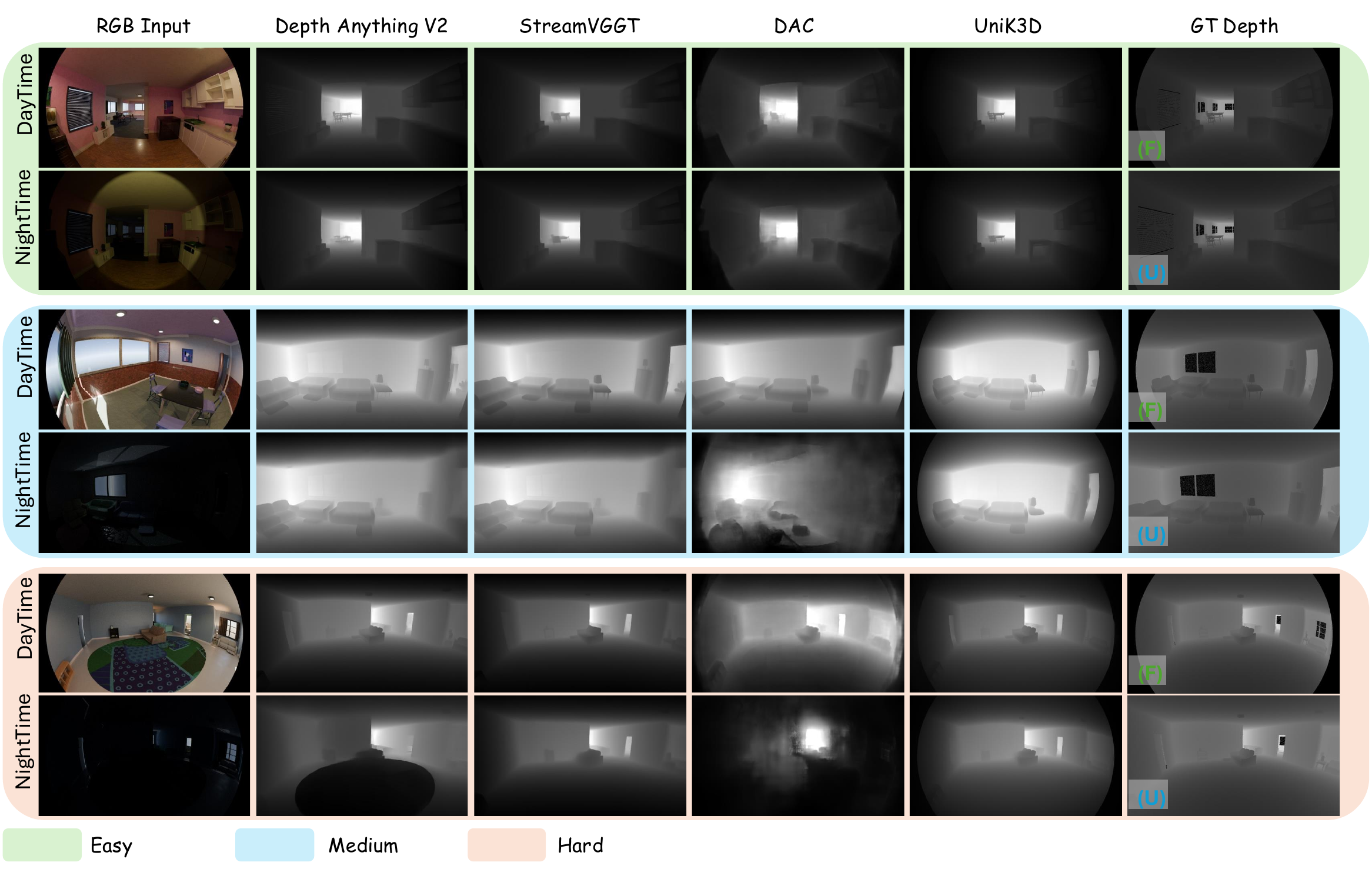}
    \caption{
     Qualitative results of monodepth estimation in day and night on EgoNight-Synthetic dataset according to different difficulty levels.
    }
    \label{fig:depth_synthetic}
    \vspace{-0.15in}
\end{figure}

\begin{figure}[h]
    \centering
    \includegraphics[width=\linewidth]{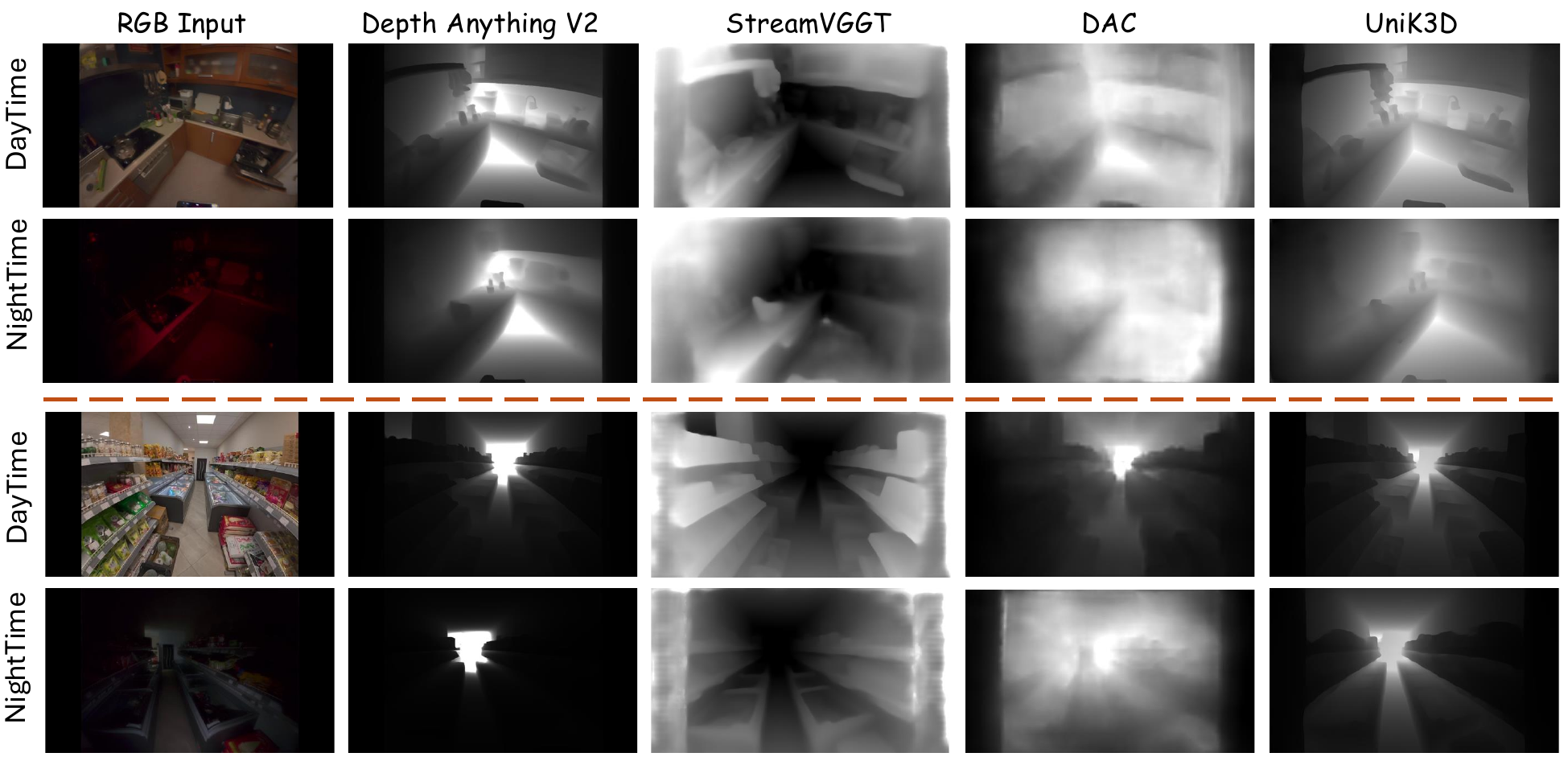}
    \caption{
     Qualitative results of monodepth estimation in day and night on EgoNight-Sofia dataset indoor part.
    }
    \label{fig:depth_sofia_indoor}
\end{figure}

\begin{figure}[h]
    \centering
    \includegraphics[width=\linewidth]{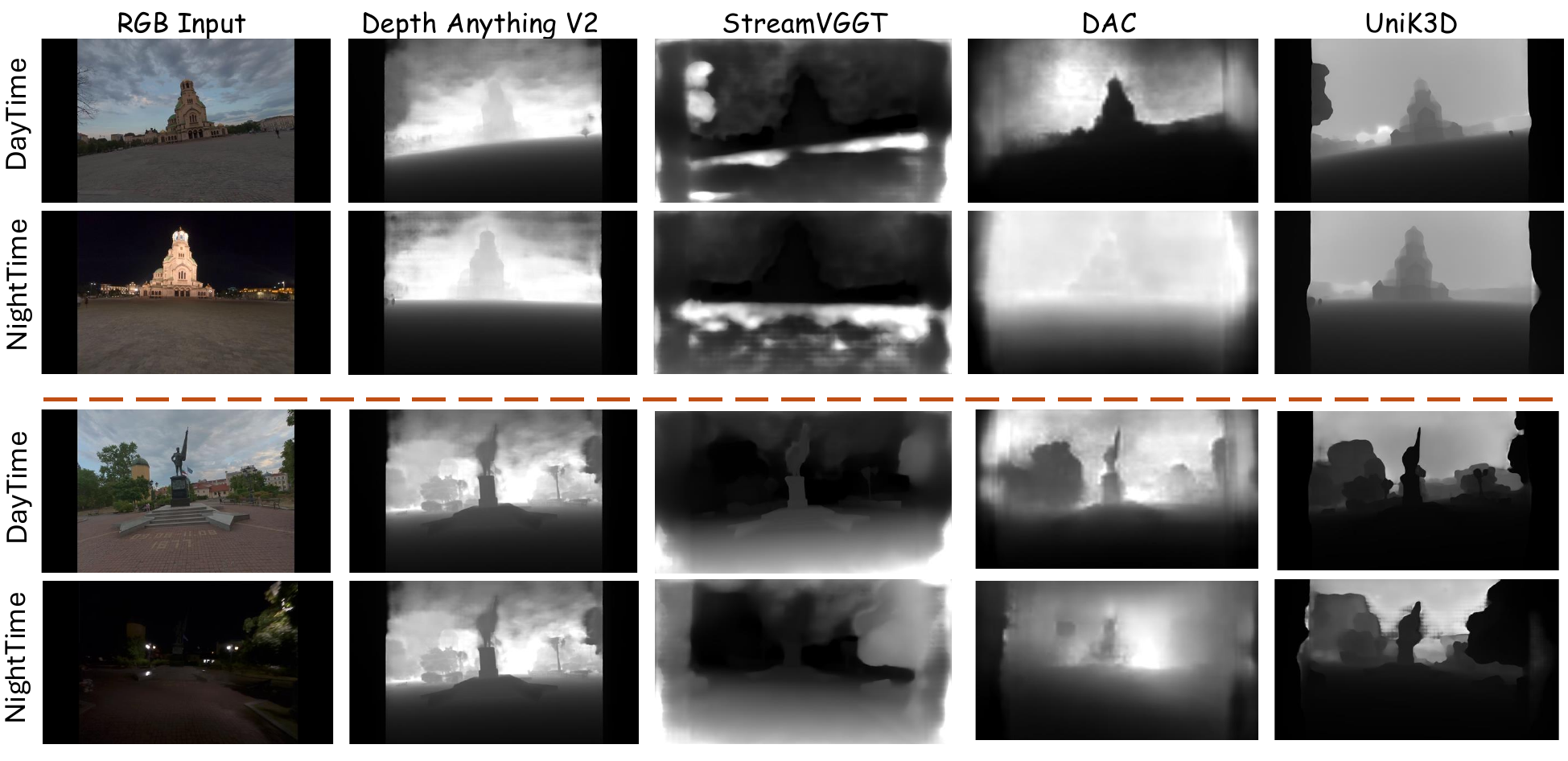}
    \caption{
     Qualitative results of monodepth estimation in day and night on EgoNight-Sofia dataset outdoor part.
    }
    \label{fig:depth_sofia_outdoor}
\end{figure}

\begin{figure}[h]
    \centering
    \includegraphics[width=.9\linewidth]{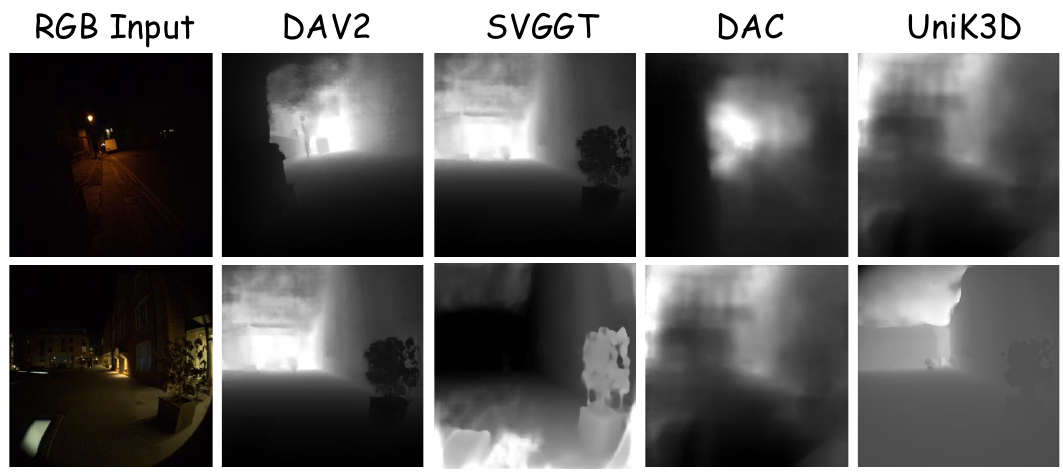}
    \caption{
     Qualitative results of monodepth estimation in day and night on EgoNight-Oxford dataset, note that DAV2 and SVGGT are shortened for Depth Anything V2 and StreamVGGT respectively.
    }
    \label{fig:depth_oxford}
\end{figure}

We provide additional qualitative results across day–night conditions (Figs.~\ref{fig:depth_synthetic}, \ref{fig:depth_sofia_indoor}, \ref{fig:depth_sofia_outdoor}, \ref{fig:depth_oxford}). Consistent with the main paper, nighttime is substantially more challenging: low SNR, head-motion blur, extreme dynamic range, color/white-balance shifts, and auto-exposure fluctuations amplify scale ambiguity and erode edge fidelity, leading to over-smoothed surfaces, depth collapse in dark regions, halos around bright point sources, and temporal instability. UniK3D remains the strongest overall in preserving scene structure under these conditions, though performance still degrades under extreme darkness and sparse texture. By contrast, \textit{StreamVGGT} and \textit{DAC} are notably brittle at night, frequently washing out structure, misinterpreting specular highlights, and producing flattened or unstable depth in large low-illumination areas. The effect is most pronounced outdoors in EgoNight-Sofia and EgoNight-Oxford, where wide dynamic range, sparse texture, and point-light saturation further depress accuracy across methods.

\subsection{More Analysis}
\subsubsection{Failurecase Analysis}
\label{supp:failure_case}
We further investigate failure reasons by systematically altering illumination, motion blur, and camera noise (sampling quality) in synthetic videos, while keeping scene, annotations, and trajectory fixed in the synthetic dataset.  The procedure of generating such difficulty level is described in Appendix.~\ref{sec:more_construction}, we calculate the averaged accuracy from Tab.~\ref{tab:main_result} as $18.47\%$ for Easy (moderately dark), $15.88\%$ for Medium (very dark with camera noise), $12.95\%$ for Hard (as dark as medium, with motion blur).  From the results, we highlight that: ii) Lower illumination together with camera sensor noise leads to the most significant drop due to loss of contrast and missing fine details. ii) Motion blur further harms performance by causing temporal ambiguity and object shape distortion. iii) These results confirm that night conditions VQA is challenging, highlighting the need for specialized night-egocentric benchmarks. We also visualize more failure cases, and give  more analysis in Appendix.~\ref{supp:more_visualization}.
\subsubsection{Limitations and future works}
\label{supp:future_work}
We acknowledge limitations of EgoNight and insights for future research directions in night video understanding.  
(1) The dataset scale remains modest compared to large-scale vision–language corpora. 
However, as a testbed, we argue that the current scale of 3,600+ human-verified QA pairs 
is already sufficient for benchmarking. In future work, we plan to further scale up nighttime 
videos by synthesizing more data and recording additional real-world footage, which will 
enable not only benchmarking but also pretraining and fine-tuning to improve MLLM performance.  

(2) We show in the main content that fine-tuning on synthetic data improves real world performance. Therefore, we can scale-up synthetic data to build a training set, that can be used to fine-tune the model, thus generalize to real world scenario. 

(3) We encourage the community to explore broader research avenues, such as:
\begin{itemize}
    \item Leveraging unlabeled nighttime or partially annotated video corpora, even if not strictly egocentric;
    
    \item Integrating low-level vision techniques for illumination enhancement or robust feature extraction;
    
    \item Exploring multimodal signal, e.g., depth from EgoNight-Synthetic, to improve low-light understanding;
    
    \item Developing training-data-free or lightweight adaptation methods that are more generalizable across MLLMs.
\end{itemize}
(4) EgoNight primarily focuses on day–night illumination shifts, while other real-world 
challenges such as weather variations (rain, fog) and extreme camera motion are not covered. 
We view these as promising directions for future extensions of EgoNight.

\subsubsection{Contribution to the Community}
\label{supp:contribution}
We believe EgoNight will serve as a valuable resource for the research community in several ways. 
First, it provides the \textit{first benchmark suite} dedicated to egocentric nighttime vision, a 
long-overlooked but practically critical setting for robust AI assistants. Second, the dataset’s 
unique day–night alignment enables rigorous analysis of illumination effects, offering insights 
that cannot be obtained from prior egocentric benchmarks. Third, by covering multiple tasks, VQA, 
day-night correspondence retrieval, and depth estimation, EgoNight provides a comprehensive testbed that can 
catalyze progress across both perception and reasoning. Finally, with all data, annotations, and 
evaluation code to be released publicly, EgoNight is designed to be easily accessible, extensible, 
and reproducible, supporting future research on egocentric vision understanding learning.

\subsubsection{Usage of Large Language Models (LLMs)}
Our annotation pipeline and benchmark evaluation both leverage large language models (LLMs). 
For data construction, advanced multimodal LLMs are used to generate initial captions, questions, 
and pseudo answers, which are then refined by human annotators. This hybrid model–human approach substantially reduces annotation cost while ensuring quality. 
For evaluation, we adopt the \textit{LLM-as-a-Judge} paradigm to assess the semantic correctness 
of model outputs against ground-truth answers, following recent practice in egocentric VQA. 
Beyond annotation and evaluation, we also used LLMs to support paper preparation, such as generating icons for illustration figures and assisting with proof-reading. Importantly, while LLMs serve as practical tools throughout our workflow, all core ideas, dataset design, experiments, and analyses are conceived and conducted 
independently by the authors.

\subsubsection{Ethic Statement}
All indoor egocentric recordings in EgoNight-Sofia were collected with explicit informed consent, and outdoor data is fully anonymized by blurring faces, license plates, house numbers, and other identifiable details, with audio removed, in compliance with GDPR and privacy standards. Before release, all videos will be verified to contain no personally identifiable information. For EgoNight-Oxford, the subset is derived from the publicly available Oxford Day-and-Night dataset under the BSD-3-Clause license, which permits redistribution and modification with proper attribution. We will retain all required license notices and appropriately acknowledge the original dataset.


\end{document}